\documentclass[dvipsnames]{article} %
\usepackage{colm2024_conference}

\usepackage{booktabs}
\usepackage{enumitem}
\usepackage{wrapfig}
\usepackage{algorithm}
\usepackage{algpseudocode}
\usepackage{graphicx}
\usepackage{subfigure}
\usepackage[misc]{ifsym}

\usepackage{microtype}
\usepackage{amsmath}
\usepackage{colortbl}
\usepackage[utf8]{inputenc}
\usepackage[T1]{fontenc}
\definecolor{lightgray}{rgb}{0.9,0.9,0.9}
\usepackage{caption}
\usepackage{subcaption}
\usepackage{setspace}
\usepackage{url}
\usepackage{multirow}
\usepackage{tabularx}
\usepackage{blindtext}
\usepackage{pgfplots}
\pgfplotsset{compat=1.18} 
\usepackage{tikz}
\usetikzlibrary{er,positioning,bayesnet}
\usepackage{makecell}
\usepackage{tipa}
\usepackage{siunitx}
\usepackage{nicefrac}
\usepackage{listings}
\usepackage[raster,skins, most]{tcolorbox} %
\usepackage{xltabular}
\usepackage{adjustbox}
\usepackage{xurl}
\usepackage{rotating}
\usepackage[normalem]{ulem}

\usepackage{fontawesome}

\useunder{\uline}{\ul}{}

\usepackage{amsmath,amsfonts,bm}

\def\eqref#1{equation~\ref{#1}}

\def\1{\bm{1}}

\DeclareMathAlphabet{\mathsfit}{\encodingdefault}{\sfdefault}{m}{sl}
\SetMathAlphabet{\mathsfit}{bold}{\encodingdefault}{\sfdefault}{bx}{n}

\renewcommand{\ghlink}{https://github.com/Alibaba-NLP/WebAgent}

\newcommand*\justify{%
  \fontdimen2\font=0.4em
  \fontdimen3\font=0.2em
  \fontdimen4\font=0.1em
  \fontdimen7\font=0.1em
  \hyphenchar\font=`\-
}

\renewcommand{\texttt}[1]{%
  \begingroup
  \ttfamily
  \begingroup\lccode`~=`/\lowercase{\endgroup\def~}{/\discretionary{}{}{}}%
  \begingroup\lccode`~=`[\lowercase{\endgroup\def~}{[\discretionary{}{}{}}%
  \begingroup\lccode`~=`.\lowercase{\endgroup\def~}{.\discretionary{}{}{}}%
  \catcode`/=\active\catcode`[=\active\catcode`.=\active
  \justify\scantokens{#1\noexpand}%
  \endgroup
}

\usepackage{makecell}
\usetikzlibrary{tikzmark}
\makeatletter
\newcommand*\myfontsize{%
  \@setfontsize\myfontsize{7}{8}%
}
\makeatother

\definecolor{uclablue}{RGB}{159, 195, 224}

\definecolor{uclagold}{RGB}{255, 240, 180}

\definecolor{aliceblue}{RGB}{255, 238, 241}

\newcommand{\mytextbox}[2]{\tikzmarknode[draw=#1,thick,inner sep=2pt]{test}{\myfontsize #2}}

\definecolor{cadmiumgreen}{rgb}{0.0, 0.42, 0.24}

\definecolor{myred}{rgb}{0.7, 0.3, 0.0}
\definecolor{myblue}{rgb}{0.2, 0.3, 0.6}
\definecolor{babygreen}{rgb}{0.85, 0.97, 0.85}

\definecolor{purple1}{RGB}{126, 107, 196}
\definecolor{purple2}{RGB}{199, 158, 207}
\definecolor{purple3}{RGB}{214, 200, 255}
\definecolor{purple4}{RGB}{254, 240, 255}

\definecolor{deepblue}{RGB}{48, 58, 82}

\newcommand{\thinkl}{\mytextbox{deepblue}{\textbf{\textcolor{deepblue}{<think>}}}}

\newcommand{\thinkr}{\mytextbox{deepblue}{\textbf{\textcolor{deepblue}{</think>}}}}

\newcommand{\calll}{\mytextbox{deepblue}{\textbf{\textcolor{deepblue}{<tool\_call>}}}}

\newcommand{\callr}{\mytextbox{deepblue}{\textbf{\textcolor{deepblue}{</tool\_call>}}}}

\newcommand{\responsel}{\mytextbox{deepblue}{\textbf{\textcolor{deepblue}{<tool\_response>}}}}

\newcommand{\responser}{\mytextbox{deepblue}{\textbf{\textcolor{deepblue}{</tool\_response>}}}}

\newcommand{\answerl}{\mytextbox{deepblue}{\textbf{\textcolor{deepblue}{<answer>}}}}

\newcommand{\answerr}{\mytextbox{deepblue}{\textbf{\textcolor{deepblue}{</answer>}}}}

\newcommand{\symboletongyi}{\raisebox{0pt}{~\includegraphics[scale=0.012]{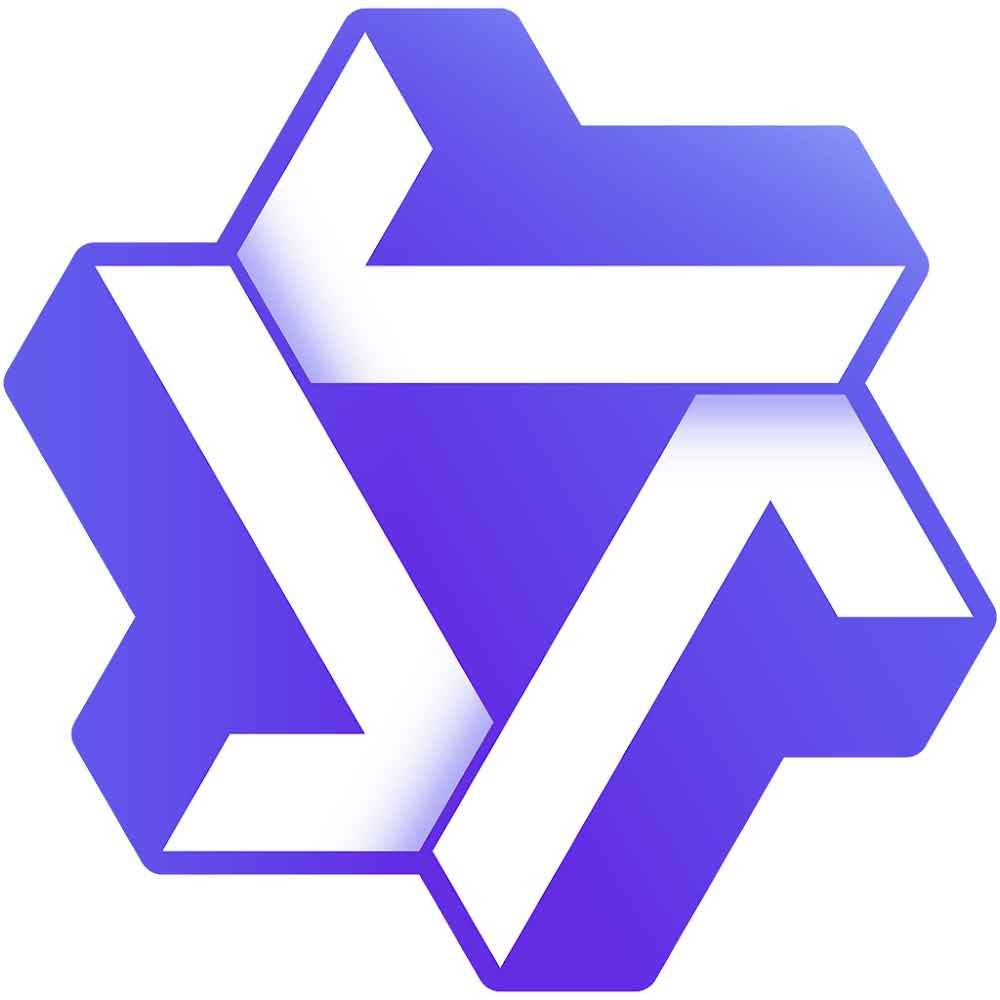}}~}

\definecolor{deepPurple}{HTML}{330066}

\definecolor{uclablue_old}{rgb}{0.15, 0.45, 0.68}
\hypersetup{
    breaklinks,
    citecolor=uclablue_old,
    colorlinks=true,
}

\newtcolorbox{mybox}[2][]
  {colback = black!5!white, colframe = black!75!black, fonttitle = \bfseries,
    colbacktitle = black!100!black, enhanced, before upper={\fontsize{8}{11}\obeyspaces\obeylines\selectfont}, fontupper=\selectfont,
    attach boxed title to top left={yshift=-2.2mm,xshift=4mm},
    title=#2,#1}

\title{WebSailor: Navigating Super-human Reasoning for Web Agent}

\author{%
\small{Kuan Li$^{*}$, Zhongwang Zhang$^{*}$, Huifeng Yin$^{*}$$^{(\textrm{\Letter})}$, Liwen Zhang$^{*}$, Litu Ou\thanks{Equal Core Contributors. Kuan Li, Zhongwang Zhang, and Huifeng Yin are project leaders.}\hspace{0.5mm}, Jialong Wu, Wenbiao Yin, Baixuan Li, Zhengwei Tao, Xinyu Wang, Weizhou Shen, Junkai Zhang, Dingchu Zhang, Xixi Wu, Yong Jiang$^{(\textrm{\Letter})}$, Ming Yan, Pengjun Xie, Fei Huang, Jingren Zhou}%
  \\[1em]               
  {\fontsize{10pt}{11pt}\selectfont          
Tongyi Lab\symboletongyi, Alibaba Group}\\
}

\begin{document}

\maketitle

\begingroup
  \renewcommand\thefootnote{\Letter}  
  \footnotetext{Corresponding author. \{yinhuifeng.yhf, yongjiang.yj\}@alibaba-inc.com} 
\endgroup

\begin{abstract}

Transcending human cognitive limitations represents a critical frontier in LLM training. Proprietary agentic systems like DeepResearch have demonstrated superhuman capabilities on extremely complex information-seeking benchmarks such as BrowseComp, a feat previously unattainable. We posit that their success hinges on a sophisticated reasoning pattern absent in open-source models: the ability to systematically reduce extreme uncertainty when navigating vast information landscapes. Based on this insight, we introduce WebSailor, a complete post-training methodology designed to instill this crucial capability. Our approach involves generating novel, high-uncertainty tasks through structured sampling and information obfuscation, RFT cold start, and an efficient agentic RL training algorithm, Duplicating Sampling Policy Optimization (DUPO). With this integrated pipeline, WebSailor significantly outperforms all open-source agents in complex information-seeking tasks, matching proprietary agents' performance and closing the capability gap.

\end{abstract}


\begin{figure}[h]
    \centering
    \includegraphics[width=1.0\linewidth]{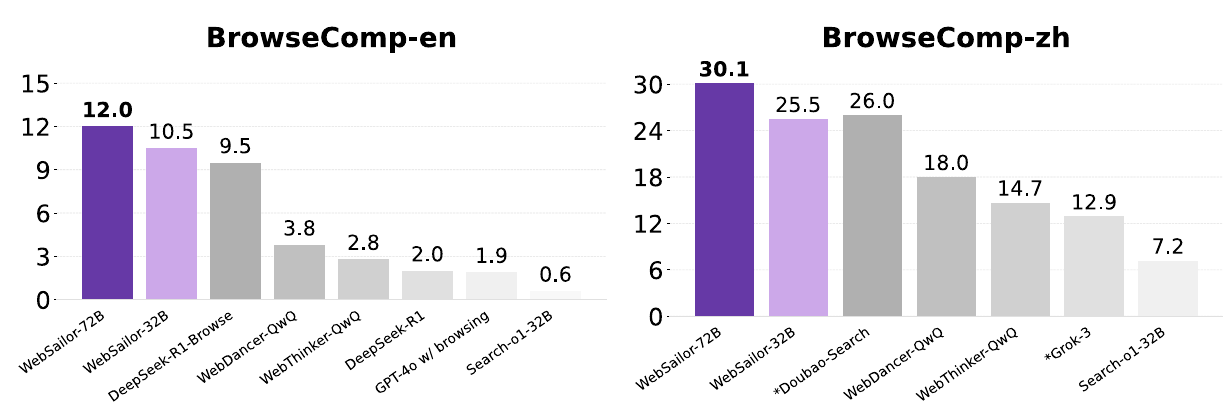}
    \caption{Performance on the BrowseComp-en/zh benchmarks. DeepSeek-R1-Browse is DeepSeek-R1 equipped with browsing tools via the ReAct framework, sharing the same implementation as our model, WebSailor. Doubao-Search and Grok-3 are proprietary web-based products (marked by $*$). The result for GPT-4o with browsing is taken from OpenAI's official publication.}
    \label{fig:abs_fig}
\end{figure}
\vfill

\newpage

\newpage
\section{Introduction}
\label{sec:intro}

Information seeking, the fundamental human drive to resolve uncertainty, has been revolutionized by the internet~\citep{wilson1999models, jurado2015measuring}.
Yet, human ability to navigate this vast digital landscape is constrained by cognitive limits: finite memory, fragile attention, and an inability to pursue multiple exploratory paths in parallel. 
Leading proprietary agentic systems, such as Deep Research~\citep{dr}, show that Large Language Model (LLM) agents can transcend these human limitations. Their superhuman performance on complex web benchmarks like BrowseComp-en/zh~\citep{bc_en, bc_zh} stems from sophisticated reasoning—internal or tool-mediated—that systematically reduces uncertainty~\citep{kapoor2024large, huang2023look}.

However, instilling these advanced reasoning capabilities in open-source agents remains an unsolved problem. As shown in Fig.~\ref{fig:abs_fig}, existing open-source LLMs and web agents exhibit near-zero accuracy on BrowseComp-en~\citep{wu2025webdancer, Li2025webthinker, li2025search, song2025r1}. This stark performance gap arises because current training paradigms focus on what we classify as Level 1 and 2 tasks: problems with either low uncertainty (e.g., single-search) or a clear, structured path to resolution (e.g., standard multi-hop QA). These datasets do not expose models to the Level 3 challenges that dominate complex benchmarks—scenarios demanding robust compositional generalization~\citep{wiedemer2023compositional} over intricate information landscapes with no predefined solution path. Consequently, models fail to develop the complex, multi-step reasoning required to navigate them.

To elicit these superhuman reasoning patterns, we generate training data characterized by high and hard-to-reduce intrinsic uncertainty. Our primary mechanism involves sampling subgraphs from interconnected knowledge structures generated by random walks across real-world websites. From a compositional generalization perspective~\citep{compge}, these subgraphs present novel combinations of known entities and relationships, forcing the model to reason about previously unseen compositions and pushing it beyond simple heuristics. This process generates a diverse array of intricate, emergent structures that are difficult to pre-define, compelling the model to develop reasoning processes that may transcend established human patterns.

We further amplify task difficulty using carefully designed information obfuscation techniques, which directly increase initial  ambiguity. The combination of structural complexity and informational ambiguity creates tasks that demand exceptionally sophisticated reasoning. For instance, some of our generated questions are so challenging that even powerful proprietary models like o3~\citep{o3} require up to 40 tool calls to arrive at a solution, underscoring the extreme uncertainty reduction involved.

After obtaining QAs, a key challenge is acquiring full supervision. While powerful open-source Large Reasoning Models (LRMs) like QwQ~\citep{qwq32b} and DeepSeek-R1~\citep{r1} can solve some complex QAs, their native reasoning outputs are unsuitable for direct fine-tuning. These models exhibit highly stylized and verbose thought processes that, if imitated, could restrict the trainee agent's ability to develop its own flexible, exploratory strategies. Furthermore, in long-horizon web tasks requiring dozens of tool calls~\cite{li2025lara}, their lengthy reasoning chains quickly overwhelm the context window, leading to performance degradation and poor readability~\citep{yin2025towards}. To overcome this, we propose a novel approach: we leverage these open-source LRMs to generate successful action-observation traces, but then reconstruct the reasoning. By inferring concise, action-oriented thoughts for each step, we create a clean, effective supervision signal that captures the solution logic without inheriting stylistic or verbosity-related drawbacks.

In terms of training process optimization, although recent studies suggest skipping SFT~\citep{r1, chen2025sft, hu2025open}, we demonstrate that a modest rejection sampling fine-tuning (RFT) cold start is indispensable for web agents navigating such complex tasks. On one hand, RL rewards for these scenarios are extremely sparse, often yielding near-zero feedback initially. On the other hand, our approach does not heavily rely on distillation; a minimal cold start with just over $2k$ high-quality examples proves effective. The RL training of agents for such tasks is extremely slow due to multi-turn reasoning and heavy tool use. To address this, we propose Duplicating Sampling Policy Optimization (DUPO), which incorporates two dynamic sampling strategies—one before training and one during training—to improve both effectiveness and efficiency.

Our family of WebSailor models (3B, 7B, 32B, and 72B) outperform all open-source models and agentic methods on BrowseComp-en/zh, and also surpass proprietary LRMs such as Grok-3~\citep{grok} and DouBao~\citep{doubao} when they are combined with browsing capabilities, as shown in Fig.~\ref{fig:abs_fig}. Additionally, we find that post-training based on complex, uncertainty-driven reasoning patterns exhibits downward compatibility, achieving promising performance on simpler tasks such as GAIA~\citep{mialon2023gaia}, XBench-DeepSearch~\citep{xbench}, and SimpleQA~\citep{simpleqa}.
\section{Problem Definition}
\label{sec:def}

We adopt the ReAct~\citep{yao2023react} as the agent's framework. Upon receiving a question, the agent performs several iterations of Thought-Action-Observation. Specifically, in each iteration, based on the existing context, the LLM generates a Thought and executes a parsable Action (tool call), then awaits the environment to return an Observation. In WebTraverseX, the action space consists of generating final answer and two tools, search and visit, which correspond to invoking a search engine with several queries and accessing several webpages via URLs to retrieve their content, respectively. The details of these two tools are provided in the Appendix~\ref{apx:tool} The observation returned by the search action consists of 10 titles, snippets, and their corresponding URLs for each search query. In contrast, the observation of the visit action is a summary of the webpages, tailored to the "goal" specified in the LLM's action. The iteration terminates when the LLM selects "final answer" as the action. A complete trajectory with $T$ iterations can be defined as:

\begin{align}
\mathcal{H}_T=(\tau_0,a_0,o_0,\dots, \tau_i,a_i,o_i, \dots,\tau_{T},a_{T}),
\end{align}

where $\tau_i$, $a_i$, $o_i$ represent thought, action, and observation in the $i$-th round, respectively. At step $t$, the thought $\tau_t$ and $a_t$ are sampled from a policy based on all previous context, i.e., $\pi(a, t|\mathcal{H}_{t-1})$. 

Completing multi-hop QA~\citep{yang2018hotpotqa, ho2020constructingmultihopqadataset} typically requires only one or two rounds of ReAct, as the actions at each step are quite clear and do not involve much strategic planning. In stark contrast, BrowseComp immerses the agent in a vast, unstructured information space where the solution path is not predefined. A naive, brute-force search is computationally infeasible, potentially requiring thousands of tool calls that would overwhelm the context window of any modern LLM. Success, therefore, hinges not on following a simple script, but on executing a highly adaptive search strategy. The agent must dynamically synthesize partial information, prune unpromising exploratory paths, and integrate disparate facts to converge on a solution. Compressing this combinatorially vast search space into a tractable trajectory of a few dozen steps requires a sophisticated chain of thought~\citep{wei2022chain}. It is precisely this process of strategic navigation and synthesis that exemplifies the complex, superhuman reasoning patterns this work seeks to elicit and model.

\section{Large-scale Training Data Synthesis for Complex Reasoning}
\label{sec:data}
In this section, we present our training data construction from two perspectives: QA construction and reasoning trajectory generation.

\subsection{SailorFog-QA: Scalable Graph-Synthesized QA}

\begin{figure}
    \centering
    \includegraphics[width=\linewidth]{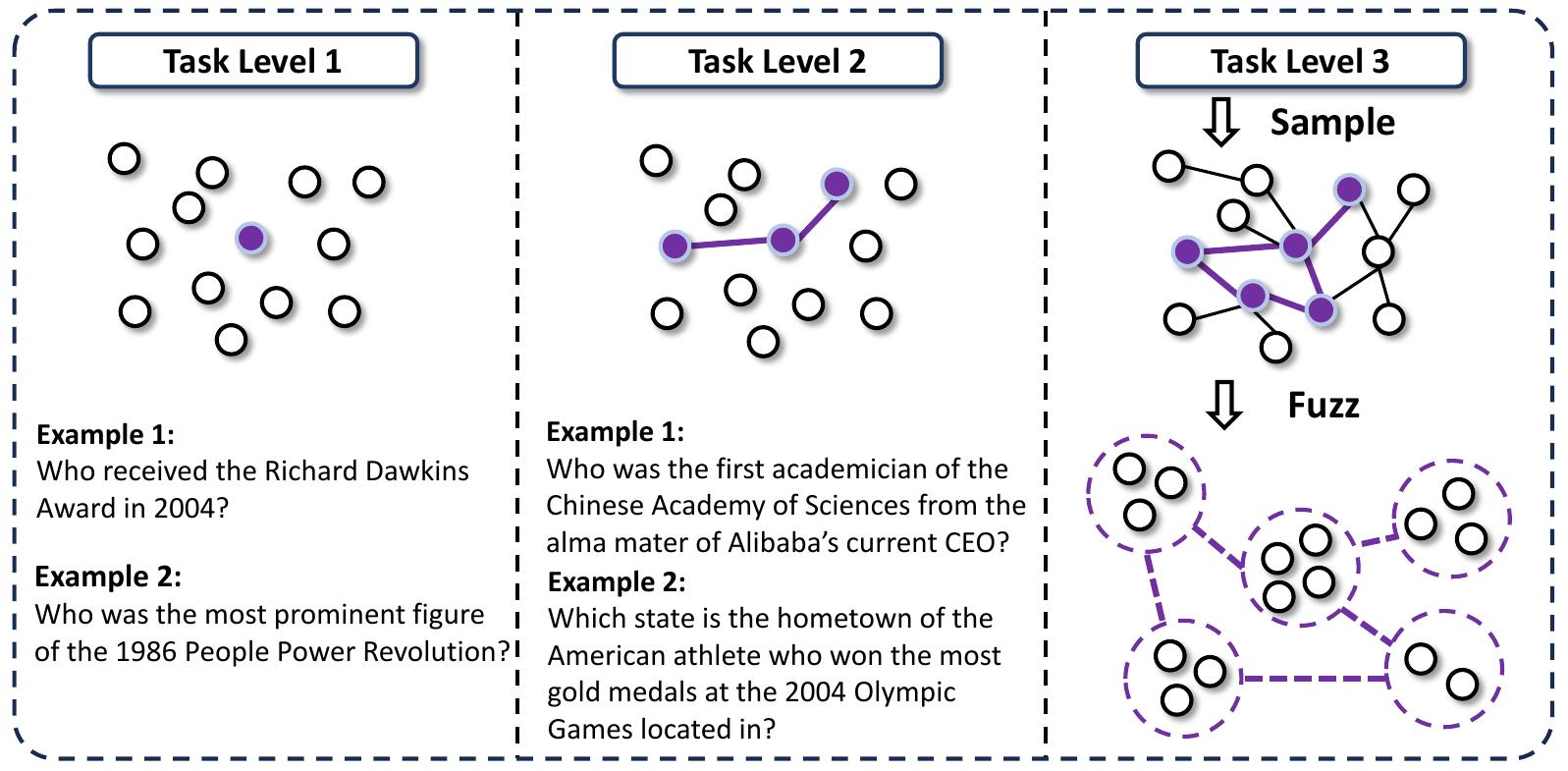}
    \caption{Information seeking tasks can be categorized into three levels. Level 1 features a relatively simple logical structure and can be answered directly or with a single, straightforward tool invocation. Level 2 resembles multi-hop questions, where solutions are obtained through a fixed sequence of reasoning steps. Level 3 exhibits the most complex and variable topology after obfuscation, making it difficult to define manually, and its inherent uncertainty is challenging to reduce.}
    \label{fig:datalevel}
\end{figure}

The reasoning patterns required to answer a question are dictated by its intrinsic uncertainty and the complexity of reducing that uncertainty. As shown in Fig. \ref{fig:datalevel}, we classify information-seeking QAs into three levels based on these two dimensions. 
\begin{itemize}
    \item \textbf{Level 1: tasks exhibit low uncertainty that is easily reduced.} These include questions answerable by the model's internal knowledge or through a single, straightforward web search.
    \item \textbf{Level 2: tasks, such as multi-hop QA, present high initial uncertainty but follow a clear path to resolution.} Even with many steps, the entities are linked by well-defined logic, allowing uncertainty to be systematically reduced through a structured sequence of actions.
    \item \textbf{Level 3: the focus of our work, involves problems with both high uncertainty and high difficulty in its reduction.} Here, entities are coupled in complex, emergent ways, lacking a pre-defined reasoning path. Solving these problems demands creative exploration and novel reasoning patterns that are difficult to specify manually.
\end{itemize}   

\paragraph{Constructing the Structural Foundation for Hard-to-Reduce Uncertainty} 
To generate Level 3 tasks, we first construct a complex informational landscape where uncertainty is inherently difficult to reduce. Our process, inspired by random walks, builds knowledge graphs with emergent, non-linear structures. We begin by seeding the graph with a fuzzy entity retrieved from Wikidata's SPARQL service to ensure a challenging starting point. Using simulated web browsing, we gather unstructured text and features about this entity from the internet. From this raw information, we extract related entities and the relationships connecting them, forming the initial nodes and edges. The crucial step is the iterative expansion: we probabilistically select existing nodes and seek out new, distinct entities to connect. This stochastic process discourages simple linear chains (characteristic of Level 2 tasks) and instead fosters a densely interconnected graph with intricate, overlapping relational paths. The resulting graph serves as a structural foundation for problems that lack a pre-defined reasoning path, compelling an agent to navigate a complex web of information rather than follow a straight line.

\paragraph{Generating High-Uncertainty Questions via Subgraph Sampling and Obfuscation}
With these complex graphs as a foundation, we generate questions characterized by high initial uncertainty. This is achieved by sampling subgraphs with diverse topologies, each representing a unique constellation of coupled entities and relations. We then formulate a question and an answer based on the subgraph. Critically, we introduce ambiguity through deliberate information obfuscation. Instead of presenting clear facts, we obfuscate features and relationships within the question. For example, a precise date is transformed into a vague period ("in the early 2010s"), a name is partially masked ("an institution founded by someone with the initial 'F'"), or a quantitative attribute is described qualitatively ("a market share of less than 1\%"). This obfuscation directly increases the initial uncertainty, forcing the agent to reason, compare, and synthesize information rather than simply executing a lookup. We name our synthetic training data \textbf{SailorFog-QA}, which has three key advantages:
\begin{itemize}
    \item The data is grounded in the real-world internet, mirroring the challenges agents face in practice.
    \item The diverse subgraph topologies naturally produce problems requiring a spectrum of complex reasoning patterns, from multi-step deduction to compositional and comparative analysis.
    \item The approach is highly scalable, as the number of potential subgraphs—and thus challenging questions—grows non-linearly with the graph size, enabling efficient large-scale data synthesis.
\end{itemize}

To illustrate the characteristics of our generated Level 3 tasks, two examples are presented below. These questions epitomize our methodology: they feature multiple, intricately coupled entities and deliberately obfuscated information, such as vague time references ("around the mid-5th century", "early 21st century") and non-specific descriptors ("a prominent South American capital", "a respected arts institution"). This combination of structural complexity and informational ambiguity creates a high degree of initial uncertainty that is exceptionally difficult to reduce. In fact, our manual evaluations confirm that these problems are intractable for human researchers under typical time constraints (e.g., within two hours), as they lack clear starting points for search and require extensive, non-linear exploration. Further details on our QA generation process are available in Appendix~\ref{apx:qa}.

\begin{mybox}[colback=gray!10]{\scriptsize{Examples of Generated Questions}}

\vspace{3pt}

\textbf{Question: \textit{There was an early Christian poetic hymn composed by a late antique writer who passed away around the mid-5th century. The year of this writer's death coincides with the last year of a scientific chronology that reconstructs environmental conditions from several centuries before the modern era. What is the name of this chronology?}}

\vspace{3pt}
\textbf{Answer: \textit{Estimated Tree-Ring Chronology: 300-450 A.D.}}

\noindent\rule{\linewidth}{0.4pt} 

\vspace{3pt}

\textbf{Question: \textit{A musical piece closely associated with a prominent South American capital features lyrics written by a notable figure who was later recognized with a distinguished local civic honor in the early 21st century. The composition’s melody was created by a musician who received formal training at a respected arts institution in western Colombia. What is the name of this musical piece?}}

\vspace{3pt}
\textbf{Answer: \textit{the Rue de Rivoli}}

\end{mybox}

\subsection{Reconstructing Reasoning from Expert LRM Trajectories}
\label{sec:reconstruct}

Having synthesized complex QA pairs, the next challenge is to generate corresponding solution trajectories for cold-start supervision. While powerful open-source LRMs like QwQ-32B~\citep{qwq32b} can provide some correct trajectories, directly using their full output for fine-tuning is counterproductive. We identify two critical issues:
\begin{itemize}
    \item \textbf{Stylistic Contamination:} These LRMs possess strong, often verbose, stylistic priors in their reasoning. Directly fine-tuning on these outputs can be overly prescriptive, stifling the agent's ability to develop its own exploratory strategies and generalize to unseen problems.
    \item \textbf{Context Overload:} The verbosity of their reasoning chains is a practical barrier for complex web agent tasks. A trajectory with dozens of tool calls can easily generate a history that exceeds the context limits, degrading performance and making the reasoning process intractable.
\end{itemize}

The process is as follows: first, we prompt an expert open-source LRM to generate a complete solution trajectory, including its native thoughts. From this full trajectory, we selectively discard the LRM's original, verbose thoughts, retaining only the successful action-observation sequence $(a_0, o_0, a_1, o_1, \dots)$. This trace represents the "what" and "how" of the solution path, but not the "why".

Next, we reconstruct the missing "why". For each step $t$ in the action trace, we possess the history up to the previous step, $\mathcal{H}_{t-1}=(\hat{\tau}_0, a_0, o_0, \dots, \hat{\tau}_{t-1}, a_{t-1}, o_{t-1})$, along with the expert's chosen action $a_t$ and the subsequent observation $o_t$. We then prompt a separate, powerful instruction-following model, $\pi^*$, to generate a new thought $\hat{\tau}_t$ that serves as a concise, logical justification for taking action $a_t$:
\begin{align}
\hat{\tau}_t \sim \pi^*(\tau|\mathcal{H}_{t-1}, a_t, o_t).
\end{align}
By iteratively applying this for every step, we synthesize a complete, high-quality reasoning trajectory $\hat{\mathcal{H}}_T=(\hat{\tau}_0,a_0,o_0, \dots, \hat{\tau}_{T},a_{T}, o_T)$ where the reasoning is clean and goal-oriented. For this reconstruction, we use another LLM and enforce a "short-CoT" style. This is a critical design choice, ensuring the final reasoning chain is compact enough for long-horizon tasks. This method allows us to scalably generate supervision data that instills complex reasoning patterns without the negative side effects of direct imitation.

\section{Reinforcement Learning with Cold Start}
\label{sec:training}

Our training methodology is a two-stage process. Inspired by recent advancements in post-training~\citep{chu2025sft, swamy2025all, ye2025limo} which highlight the efficacy of targeted fine-tuning before more complex learning, we first employ a modest RFT phase as a "cold start".
This initial phase aims to equip the model with fundamental tool-use capabilities and adherence to the long-horizon reasoning skeleton. Subsequently, we leverage RL to further refine the agent's reasoning abilities, enhance its sample efficiency~\citep{yue2025does}, and enable fuller utilization of our high-quality, complex training data.

\subsection{Rejection Sampling Fine-Tuning}

\paragraph{Setup} Within a complete trajectory $\mathcal{H}_T$, the agent's thoughts ($\tau_i$) are enclosed by \thinkl\ and \thinkr\ tags. Actions ($a_i$) are demarcated by \calll\ and \callr\ for function calls, or \answerl\ and \answerr\ for final responses. The environment's observations ($o_i$) resulting from tool calls are wrapped with \responsel\ and \responser\ tags. Different segments are separated by these special tokens.

\paragraph{Filtering} We apply a \emph{three-stage} filtering process to the expert-generated trajectories. Firstly, to guarantee the correctness of the supervisory signal, we conduct rejection sampling that only trajectories culminating in a correct final answer are retained. Secondly, acknowledging that the expert models possess superior long-context processing capabilities compared to our policy model, we discard any trajectory exceeding 32k tokens in length. Thirdly, we filter for task complexity by retaining those trajectories with more than $5$ tool calls, as intricate reasoning patterns and effective planning strategies typically manifest through a more extended sequence of decision-making steps.

\paragraph{Training objective} The training objective is to specifically enhance the agent's decision-making capability—that is, its ability to generate effective thoughts and actions. Consequently, the tokens corresponding to the environment's observations ($o_i$) are masked out from the loss calculation~\citep{chen2023fireact}.

\subsection{Duplicating Sampling Policy Optimization}
    Following the RFT cold-start phase, which equips the model with fundamental tool-use capabilities and adherence to a reasoning skeleton, we propose Duplicating Sampling Policy Optimization (DUPO) to further refine the reasoning abilities, enhance the sample efficiency~\citep{yue2025does} and ultimately elicit its intrinsic potential to discover and internalize sophisticated problem-solving strategies beyond direct imitation. 

The main difference between RL for agents and conventional reasoning tasks is that rollout is a multi-turn process involving interaction with the environment (tool responses)~\citep{sun2024llm}. However, the interaction with the environment causes the rollout speed of agent RL to be much slower compared to standard RL. DAPO~\citep{yu2025dapo} employs dynamic sampling to filter out rollouts that are entirely correct or incorrect, subsequently filling the batch to its target size with new QAs. While this is effective for data curation, it may necessitate sequential rollouts for different cases within the same batch. This sequential processing further exacerbates the slow training speeds characteristic of agentic RL.

To solve this issue, we first filter out overly simple cases (those with all $8$ rollouts correct) before training. During training, instead of using padding to expand the batch, we duplicate samples from the same batch that have a non-zero standard deviation. Compared to DAPO’s dynamic sampling, this approach achieves approximately 2–3 times speedup. Similar to SFT, it is also necessary to mask observations when calculating the policy loss~\citep{jin2025search}. We follow GPRO~\citep{shao2024deepseekmath} to estimate the advantage in a group-relative manner. We also utilize the token-level policy gradient loss and higher clip techniques in DAPO. The training objective of DUPO is defined as follows:

\begin{equation}
\begin{aligned}
\mathcal{J}(\theta) =\quad& \mathbb{E}_{(q,y)\sim \mathcal{D}, \{o_i\}_{i=1}^G\sim \pi_{\theta_\text{old}}(\cdot\mid context)}\\&
\Bigg[\frac{1}{\sum_{i=1}^{G}|o_i|}\sum_{i=1}^{G}\sum_{t=1}^{|o_i|} 
\min \Big( r_{i,t}(\theta) \hat{A}_{i,t},  
\ \text{clip} \Big( r_{i,t}(\theta), 1 - {\varepsilon_{low}}, 1 + {\varepsilon_{high}} \Big) \hat{A}_{i,t} \Big) \Bigg]
\\
\text{s.t.}\quad& 0< \Big|\{o_i\mid\textbf{is\_equivalent}(y,o_i)\}\Big|< G,
\label{eq:dapoloss}
\end{aligned}
\end{equation}
where $(q,y)$ is the question-answer pair, $r_{i,t}(\theta)$ is the importance sampling ratio, and $\hat{A}_{i,t}$ is an estimator of the advantage at time step $t$:
\begin{equation}
    r_{i,t}(\theta)=\frac{\pi_{\theta}(o_{i,t} \mid context)}{\pi_{\theta_{\text{old}}}(o_{i,t} \mid context)},\quad\hat{A}_{i,t} = \frac{R_i - \text{mean}(\{R_i\}_{i=1}^G)}{\text{std}(\{R_i\}_{i=1}^G)}.
\label{eq:advantage_calculation}
\end{equation}

Notably, in Eq.~\ref{eq:advantage_calculation}, $o_i$ represents the tokens generated by the model but not the whole trajectory. Meanwhile, \emph{context} comprises the model generation and tool response. Cases with a standard deviation of 0 (i.e., all roll-out answers are either completely correct or completely incorrect) are removed. These slots in the batch were then filled by randomly duplicating other cases within the same batch whose standard deviation was not 0.

To avoid reward hacking~\citep{amodei2016concrete, o3}, we adopt a rule-based reward that combines both format validation and answer validation:
\begin{equation}
    R_i = 0.1 * R_i^{format} + 0.9 * R_i^{answer}.
\end{equation}
Specifically, the format score verifies whether the rollout trajectory follows the predefined format, such as whether different content segments are correctly wrapped with tags like \thinkl~ and \calll, and whether the sequence complies with the ReAct framework. The answer score uses an LLM as a judge to determine whether the final prediction is correct.
\section{Experiments}

\subsection{Setup}
\paragraph{Models and Benchmarks} We perform RFT and RL training on Qwen-2.5-3B, Qwen-2.5-7B, Qwen-2.5-32B, Qwen-2.5-72B. We mainly evaluate our method on four challenging benchmarks: 
\begin{itemize}
    \item \textbf{BrowseComp-en}~\citep{bc_en}: one of the most challenging benchmarks introduced by OpenAI to evaluate the proficiency of AI agents in locating hard-to-find, often multi-faceted, information across the internet, which demands sophisticated browsing strategies and reasoning capabilities.
    \item \textbf{BrowseComp-zh}~\citep{bc_zh}: Similar to BrowseComp-en, but the QAs are in Chinese.
    \item \textbf{GAIA}~\citep{mialon2023gaia}: A benchmark that requires multi-modality and tool-use abilities. We only use a subset of 103 cases from the text-only validation subset~\citep{Li2025webthinker, wu2025webdancer}.
    \item \textbf{XbenchDeepSearch}~\citep{xbench}: A new, dynamic, professionally-aligned benchmark that focuses on evaluating AI agents' tool usage capabilities, specifically in deep information retrieval and complex search tasks.
\end{itemize}

\paragraph{Baselines} We compare our method with the following paradigms:
\begin{itemize}
    \item \textbf{Direct Inference}: Models answer questions based on its internal knowledge. For the non-reasoning model, we choose Qwen-2.5-32B, Qwen-2.5-72B~\citep{qwen2.5}, GPT-4o~\citep{gpt4o}, GPT-4.1~\citep{gpt4.1}, and for the reasoning models, we select QWQ-32B~\citep{qwq32b}, o4-mini~\citep{o3}, and DeepSeek-R1~\citep{r1}. We do not consider smaller models because their scores on BrowseComp are essentially zero.
    \item \textbf{Proprietary Browsing Agents}: We test OpenAI DeepResearch~\citep{dr}, Grok-DeepResearch \citep{grok}, and Doubao with Deep Think and Search~\citep{doubao}; however, as not all of them are fully accessible via API, they were not tested across all benchmarks and experiments.
    \item \textbf{Open-source Agents}: We compare our method with recent open-source web/search agents, including Search-o1~\citep{li2025search}, WebThinker~\citep{Li2025webthinker}, R1-Searcher~\citep{song2025r1}, and WebDancer~\citep{wu2025webdancer}.
\end{itemize}

\paragraph{Metric and Hyper-parameters} 
We default to pass@$k$ evaluation~\citep{chen2021evaluating} and report pass@1 using non-zero temperature, and temperature and top-p are set to 0.6 and 0.95. For accuracy, we use LLM as a judge~\citep{DBLP:conf/coling/LiuYHZHWDSZ24, DBLP:conf/emnlp/WangCCL0WYXZLLY24}.  The pass@1 is computed as:

\begin{equation}
\text { pass@1 }=\frac{1}{n} \sum_{i=1}^n p_i,
\end{equation}

where $p_i$ denotes the correctness of the $i$-th response. For pass@k that $k > 1$ we repeatedly generate for $k$ times.

\subsection{Main Results}

\begin{table}[h]
    \caption{Main results on four challenging benchmarks. $^\ddagger$ indicates that these proprietary methods are manually evaluated through their websites (some are reported in the corresponding benchmark papers). - means that we do not have the results due to cost constraints.} 
    
    \centering
    \resizebox{\textwidth}{!}{\begin{tabular}{lc|c|c|c|c}
    \toprule
    \textbf{Backbone} & \textbf{Paradigm} & \textbf{BrowseComp-en} & \textbf{BrowseComp-zh} & \textbf{Xbench-DeepSearch} & \textbf{GAIA} \\
    \midrule
    \rowcolor{gray!33}\multicolumn{6}{c}{\emph{\textbf{Direct Inference}}} \\
    \midrule
    
    Qwen-2.5-32B & Direct & 0.6 & 3.9 & 8.7 & 13.6\\
    Qwen-2.5-72B & Direct & 0.6 & 7.0 & 12.7 & 14.6\\
    GPT-4o & Direct & 0.6 & 6.2 & 18.0 & 17.5\\
    GPT-4.1 & Direct & 1.5 & 14.4 & 17.0 & 22.3\\
    QwQ-32B & Direct & 0.5 & 10.0 & 10.7 & 22.3 \\
    o4-mini & Direct & \textbf{6.1} & 15.2 & 22.3 & \textbf{33.3}    \\
    DeepSeek-R1 & Direct & 2.0 & \textbf{26.3} & \textbf{32.7} & 16.5\\
    
    \midrule
    \rowcolor{gray!33}\multicolumn{6}{c}{\emph{\textbf{Proprietary Agents}}} \\
    \midrule

    Grok-3$^\ddagger$ & Browsing & - & 12.9 & 50+ & -\\
    Doubao$^\ddagger$ & Browsing & - & 26.0 & 50+ & -\\
    GPT-4o$^\ddagger$ & Browsing & 1.9 & - & - & -\\ 
    DeepResearch$^\ddagger$ & Browsing & \textbf{51.5} & \textbf{42.9} & - & \textbf{67.4}\\

    \midrule
    \rowcolor{gray!33}\multicolumn{6}{c}{\emph{\textbf{Open-source Agents}}} \\
    \midrule

    R1-Searcher-7B & ReAct & 0.4 & 0.6 & 4.0 & 20.4\\
    Qwen-2.5-32B & Search-o1 & 0.1 & 2.4 & 3.7 & 28.2\\
    WebDancer-32B & ReAct & 2.5 & 14.1 & 38.7 & 40.7\\
    QwQ-32B & Search-o1 & 2.8 & 17.9 & 25.0 & 39.8\\
    WebThinker-RL & ReAct & 2.8 & 7.3 & 24.0 & 48.5\\ 
    WebDancer-QwQ & ReAct & 3.8 & 18.0 & 39.0 & 51.5\\
    \midrule
    WebSailor-3B & ReAct & 3.3 & 9.7 & 27.7 & 33.0 \\
    WebSailor-7B & ReAct & 6.7 & 14.2 & 34.3 & 37.9 \\
    WebSailor-32B & ReAct & 10.5 & 25.5 & 53.3 & 53.2\\
    WebSailor-72B & ReAct & \textbf{12.0} & \textbf{30.1} & \textbf{55.0} & \textbf{55.4}\\
    \bottomrule
    \end{tabular}}
    \label{tab:main}
\end{table}

Our main results, presented in Table~\ref{tab:main}, reveal several critical insights.
\paragraph{The Inadequacy of Direct Inference for Complex Information Seeking} Relying solely on a model's internal knowledge and reasoning capabilities is insufficient for solving complex information retrieval tasks. Across the board, all models, including strong proprietary model like GPT-4.1, exhibit poor performance on BrowseComp-en/zh, with accuracy scores often near zero. This starkly demonstrates that these tasks require dynamic interaction with an external information source—the web—to gather the necessary evidence. The inherent uncertainty and specificity of the questions far exceed the scope of pre-trained knowledge, underscoring the necessity of an agentic, tool-using framework.

\paragraph{Superior Reasoning Models Show a Glimmer of Potential} Within the direct inference paradigm, a notable exception is the superior performance of leading reasoning models like DeepSeek-R1 and o4-mini compared to other base models. For instance, DeepSeek-R1 achieves a score of 26.3 on BrowseComp-zh, significantly higher than other models in its category. This suggests that their advanced intrinsic reasoning capabilities allow them to better decompose complex questions and reduce uncertainty to some extent, even without external tools.

\paragraph{WebSailor Establishes a New State-of-the-Art for Open-Source Agents} WebSailor sets a new state-of-the-art for open-source agents, with its advantage being most pronounced on the exceptionally challenging BrowseComp-en and BrowseComp-zh benchmarks. This result validates our core hypothesis: training on data synthesized to embody complex, hard-to-reduce uncertainty endows an agent with robust and generalizable reasoning strategies. The efficacy of our approach is strikingly demonstrated by WebSailor-3B and WebSailor-7B. Despite their modest size, WebSailor-7B achieves an accuracy of 6.7 on BrowseComp-en, decisively outperforming agents built on much larger 32B models, such as WebDancer-32B (2.5) and WebThinker-RL (2.8). This underscores that the performance gains are driven by our novel training paradigm—sophisticated data synthesis and targeted reinforcement learning—rather than being a mere artifact of model scale. While WebSailor performs strongly across all benchmarks, its margin on GAIA is more modest. Our manual inspection reveals this is because a significant portion of GAIA tasks requires mathematical and computational abilities, for which WebSailor was not specifically optimized. However, its accuracy on the purely information-retrieval subsets of GAIA remains exceptionally high, reaffirming its specialized expertise.

\paragraph{Achieving Parity with Proprietary Systems} Perhaps the most significant finding is that WebSailor closes the gap between open-source and leading proprietary systems. On BrowseComp-zh, WebSailor-72B achieves performance on par with Doubao, a top-tier proprietary agent. While the SOTA system DeepResearch still holds a lead, WebSailor's performance represents a major milestone, demonstrating that with sophisticated data synthesis and targeted training strategies like DUPO, open-source models can be elevated to a level of capability previously exclusive to closed, proprietary systems.

\subsection{Analysis}

\begin{figure}[htbp]
    \centering
    \subfigure[Comparison with BrowseComp-en]{\includegraphics[height=0.25\textwidth]{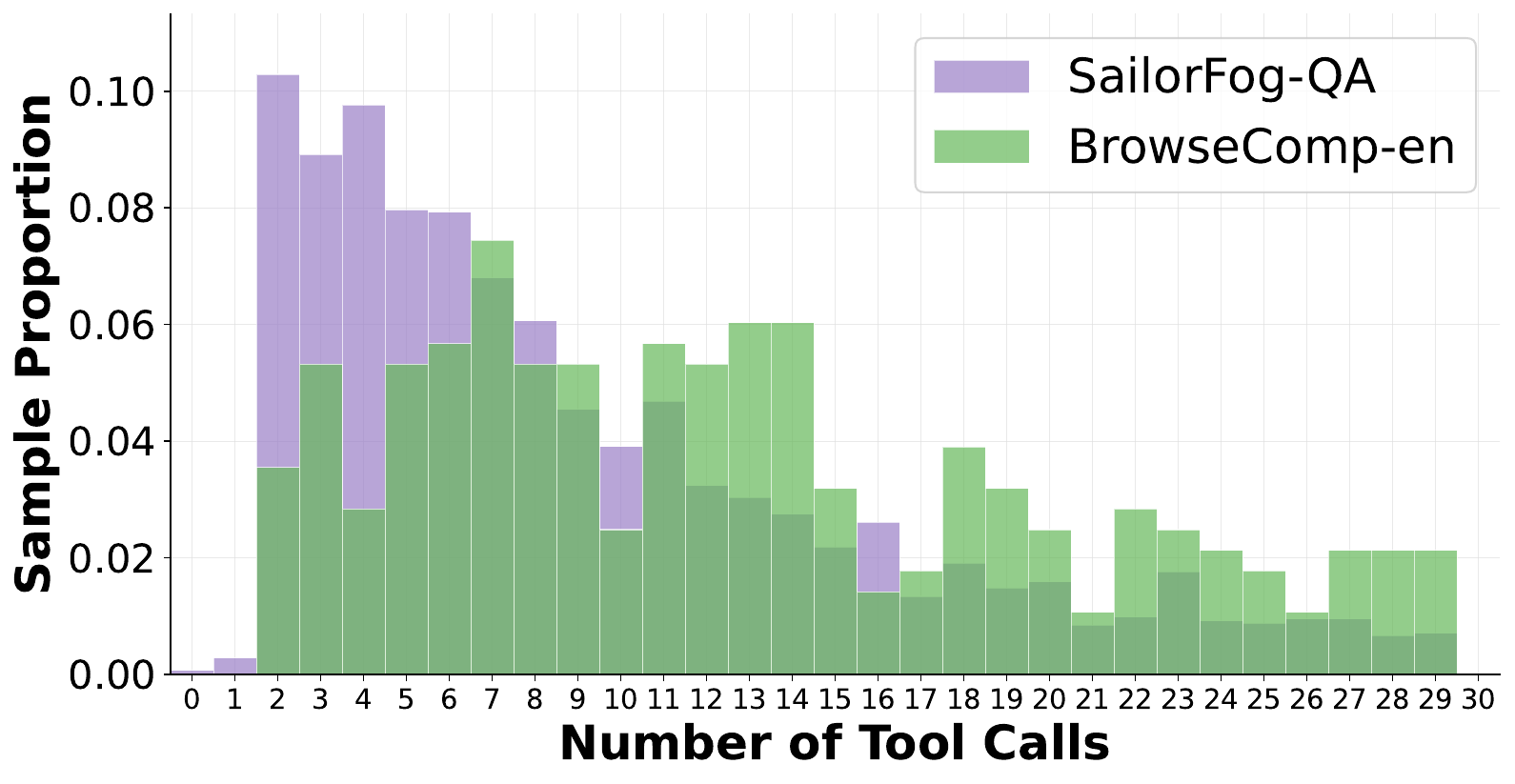}}
    \subfigure[Comparison with WebDancer]{\includegraphics[height=0.25\textwidth]{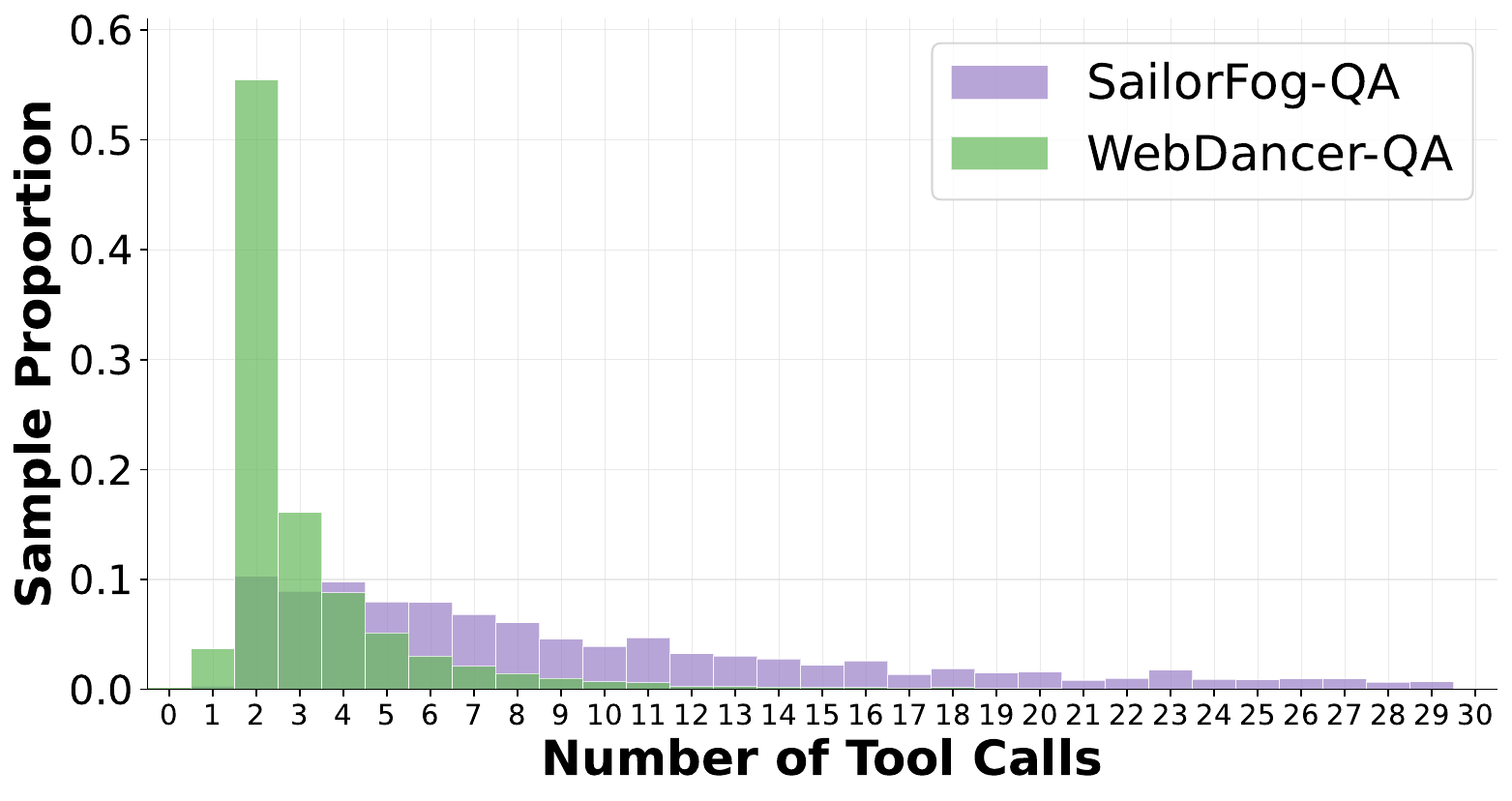}}
    \caption{A comparison of the number of tool calls in our training set with those in the training sets of WebDancer and BrowseComp-en.}
    \label{fig:tool_call}
\end{figure}

\paragraph{Complexity of SailorFog-QA} Figure \ref{fig:tool_call} provides a quantitative analysis of task complexity by plotting the distribution of tool call counts for our expert-generated training data against both the BrowseComp-en benchmark and the WebDancer training set. We use the number of tool calls as a proxy for problem difficulty. This analysis is based on unfiltered but correct trajectories from rejection sampling. The WebDancer dataset is heavily skewed towards simplicity, with over 50\% of its trajectories requiring only two tool calls and virtually none exceeding ten. In sharp contrast, our synthesized data exhibits a long-tail distribution, with a significant concentration of samples requiring more than five tool calls and extending to trajectories with over twenty interactions. Crucially, this distribution closely mirrors the complexity profile of the BrowseComp-en benchmark itself. It is important to note that the figure displays our data before our final filtering stage, where we retain only trajectories with more than five tool calls. This strategic data construction ensures that our model is trained on problems that are not only complex but also structurally representative of the hard reasoning tasks, thereby equipping it with the robust, multi-step reasoning capabilities necessary for success.

\begin{wraptable}{r}{0.5\linewidth}
\caption{The pass@1 accuracy of the SailorFog-QA, the WebDancer training set, and BrowseComp-en under the ReAct framework.}
\label{tab:rej}
\resizebox{\linewidth}{!}{
    \begin{tabular}{c|c|c|c}
    \toprule
    Backbone & SailorFog-QA & WebDancer-QA & BrowseComp-en \\
    \midrule
    o4-mini & 47.3 & 90.2 & 26.3 \\
    DeepSeek-R1 & 38.9 & 84.4 & 9.5 \\

    \bottomrule
    \end{tabular}
}
\end{wraptable} 

\paragraph{Pass rate of SailorFog-QA}To further understand the difficulty of our synthetic data, Table~\ref{tab:rej} presents the pass@1 accuracy of SailorFog-QA before filtering. DeepSeek-R1 and o4-mini are equipped with browsing tools and ReAct framework. We observe that, before filtering, our data is significantly more difficult than the WebDancer training set. Although the difficulty is lower than BrowseComp-en, it is worth noting that BrowseComp-en filters out simple cases~\citep{bc_en}. Upon manual inspection, we find that the low accuracy in our data is partly due to its inherent difficulty, but also because there may not always be a unique answer. Ambiguity in the information can result in multiple intersections of conditions that do not yield a single definitive answer—this is similar to the situation in BrowseComp-en. However, we can ensure the correctness of the conditions relative to the answer, i.e., the answer always satisfies the constraints specified in the question.

\begin{figure}[t]
    \centering
    \includegraphics[width=1\linewidth]{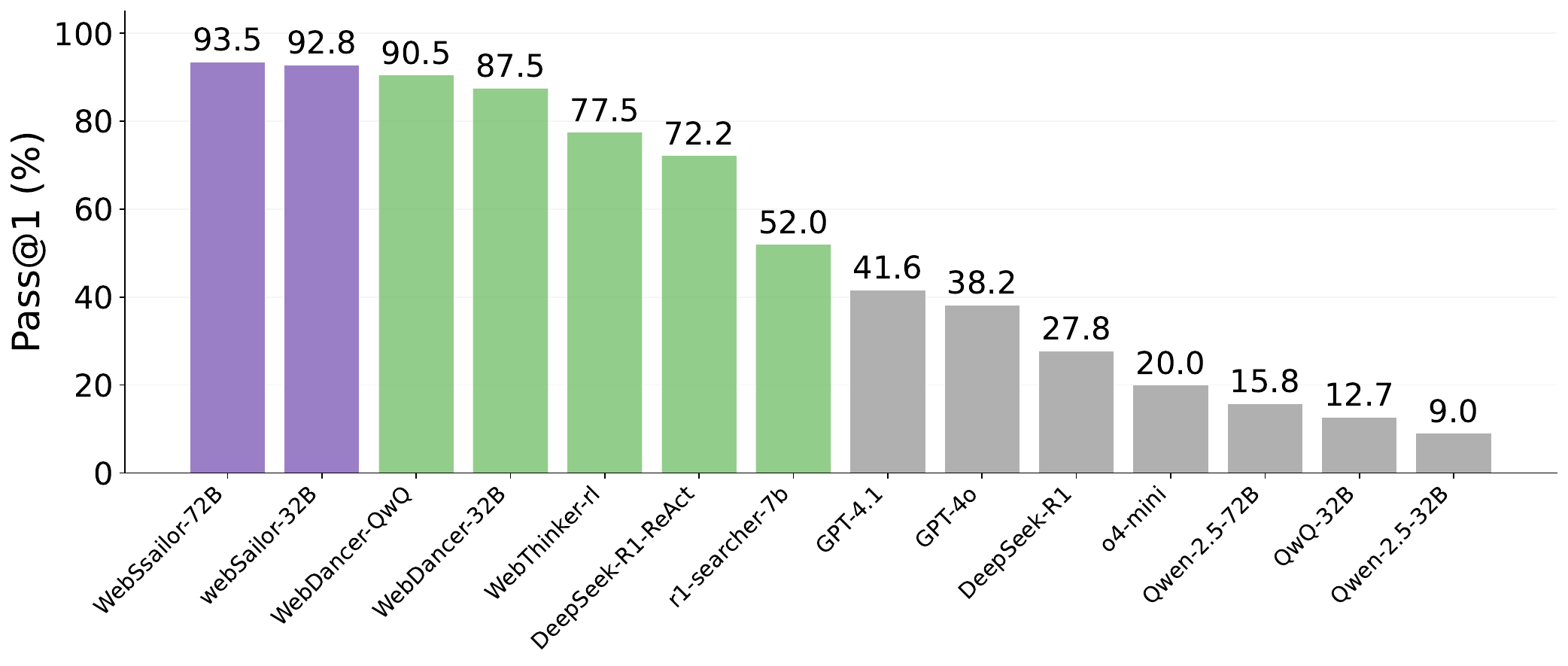}
    \caption{Performance on the SimpleQA benchmark.}
    \label{fig:simple}
\end{figure}

\paragraph{Compatibility with simple tasks} WebSailor is trained exclusively on high-difficulty data, while BrowseComp-en/zh, GAIA, and Xbench can all be considered as level-2 or level-3 tasks according to our definition. To verify whether WebSailor still performs strongly on simpler level-1 tasks, we evaluate its performance on a subset of SimpleQA benchmark~\citep{wei2024measuring}. The complete SimpleQA dataset contains 4,326 QA pairs. Since testing on the entire set would be too time-consuming, we randomly sample 200 QA pairs for evaluation. This benchmark is characterized by high correctness and fact-based questions with simple conditions, and it is challenging for frontier LLMs to answer directly. The results, as shown in Figure \ref{fig:simple}, indicate that almost all agent-based methods outperform direct answering. WebSailor surpasses all other methods, demonstrating its compatibility and effectiveness even on simpler tasks.

\begin{figure}[htbp]
    \centering
    \includegraphics[width=1\linewidth]{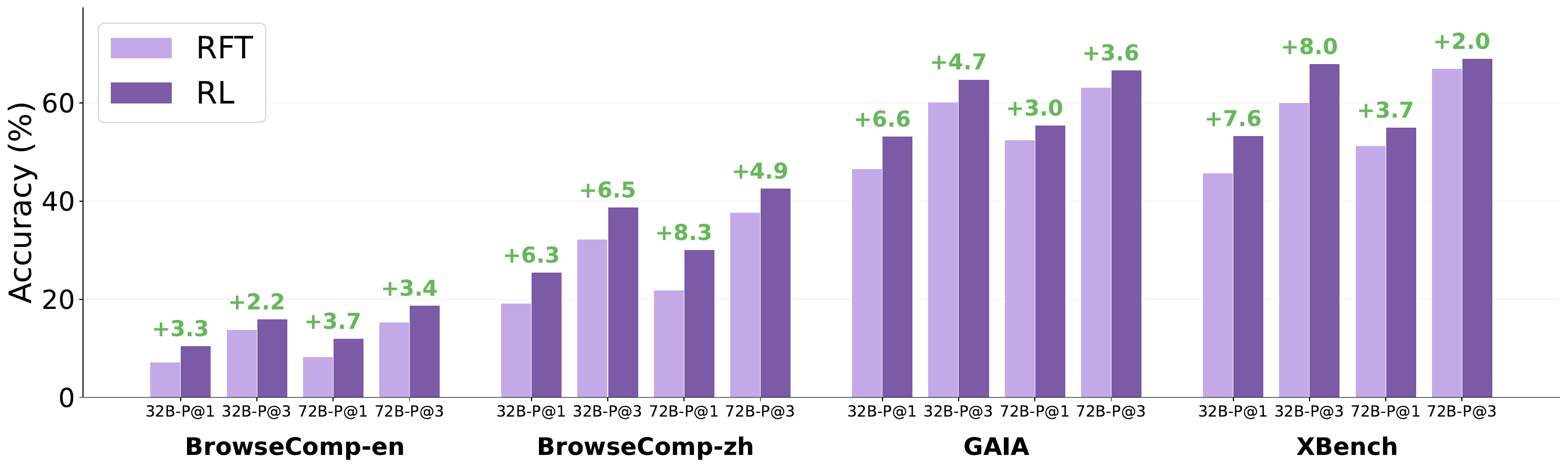}
    \caption{Detailed evaluation results using Pass@1, Pass@3.}
    \label{fig:scaling}
\end{figure}

\paragraph{Pass@1 vs Pass@3} We analyze the impact of our RL training by comparing the Pass@1 and Pass@3 performance of WebSailor before and after the RL stage (Fig.~\ref{fig:scaling}). The results reveal that RL brings notable improvements across all benchmarks, with the most significant gains observed on the highly difficult BrowseComp-en/zh tasks. This disparity is telling: the extreme complexity of BrowseComp requires agents to generate exceptionally long and intricate trajectories, making stable, repeatable success challenging~\citep{sun2025climbing}. This instability is evident in the wide initial gap between Pass@1 and Pass@3 scores for BrowseComp. RL training directly addresses this issue by reinforcing successful strategies and pruning ineffective ones, which significantly improves the model's ability to converge on a coherent solution path. Consequently, the model's stability is enhanced, leading to greater gains on harder benchmarks. Moreover, we observe that the improvement in Pass@1 is proportionally much larger than in Pass@3, indicating that RL substantially enhances sample efficiency~\citep{yue2025does}, allowing the model to achieve near its full potential with just a single sample.

\begin{figure}[t]
    \centering
    \subfigure[Pass@1 on BrowseComp-en.]{\includegraphics[height=0.22\textwidth]{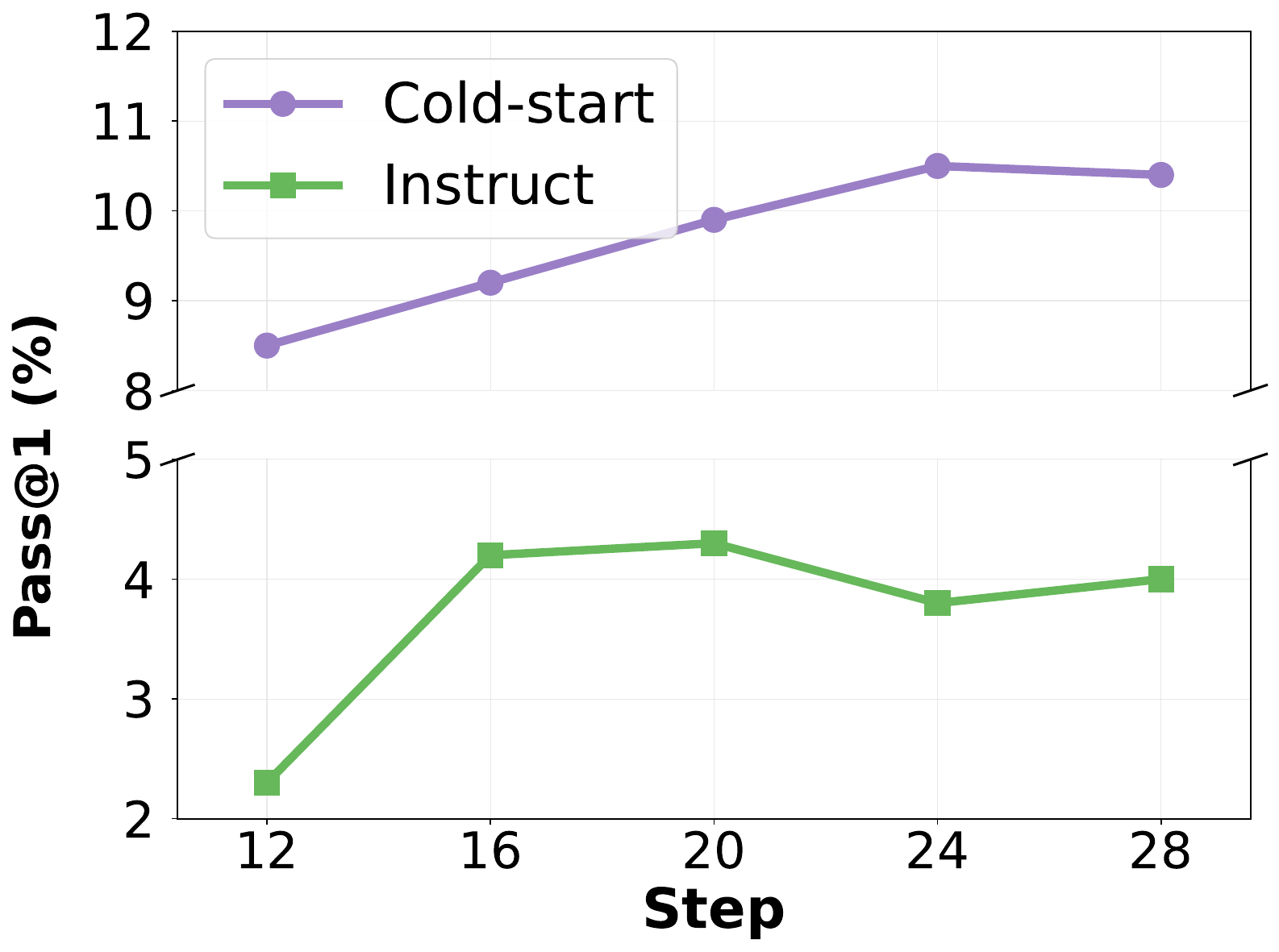}}
    \subfigure[Pass@1 on GAIA.]{\includegraphics[height=0.22\textwidth]{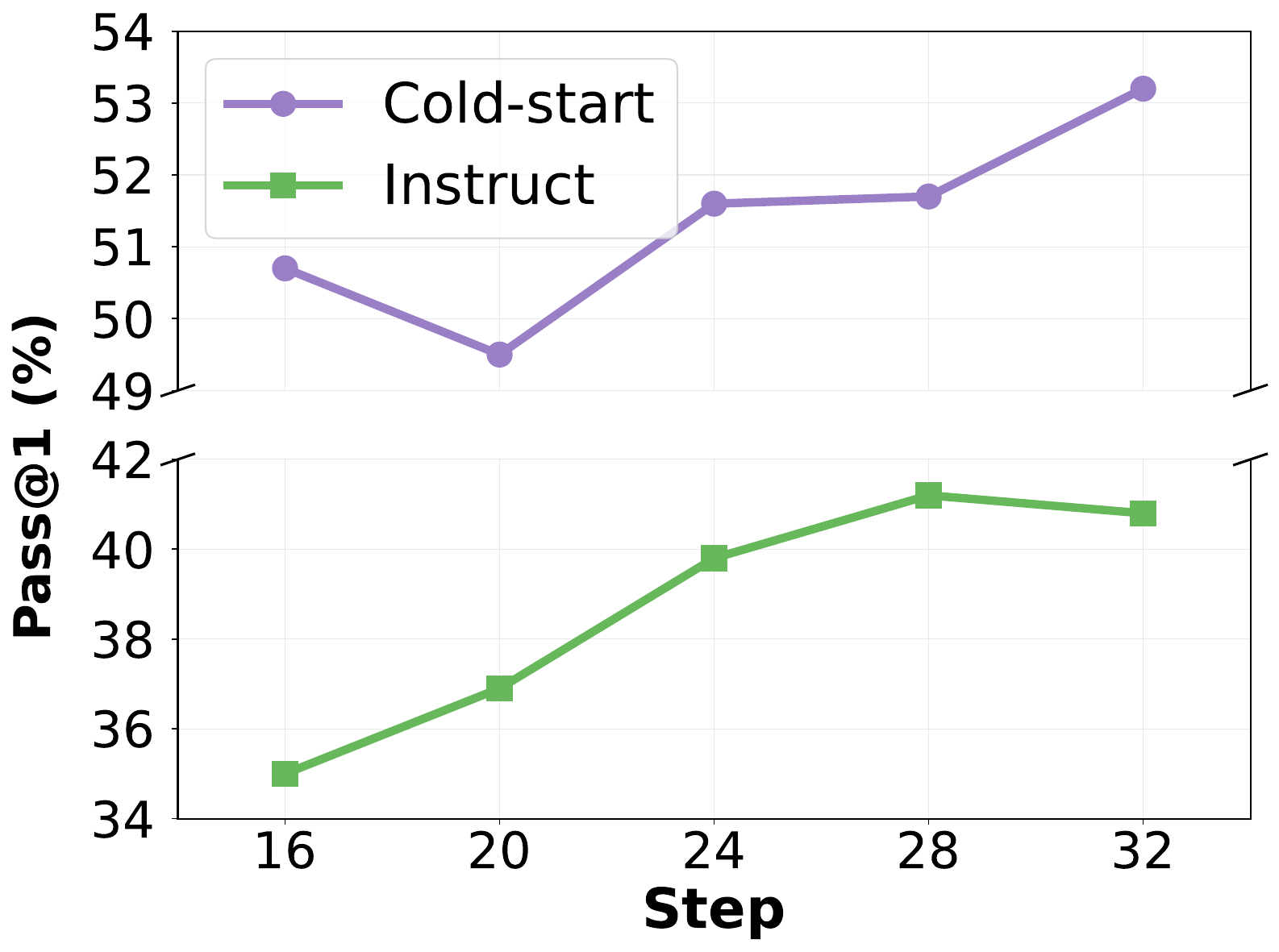}}
    \subfigure[Change in the number of tool calls.]{\includegraphics[height=0.22\textwidth]{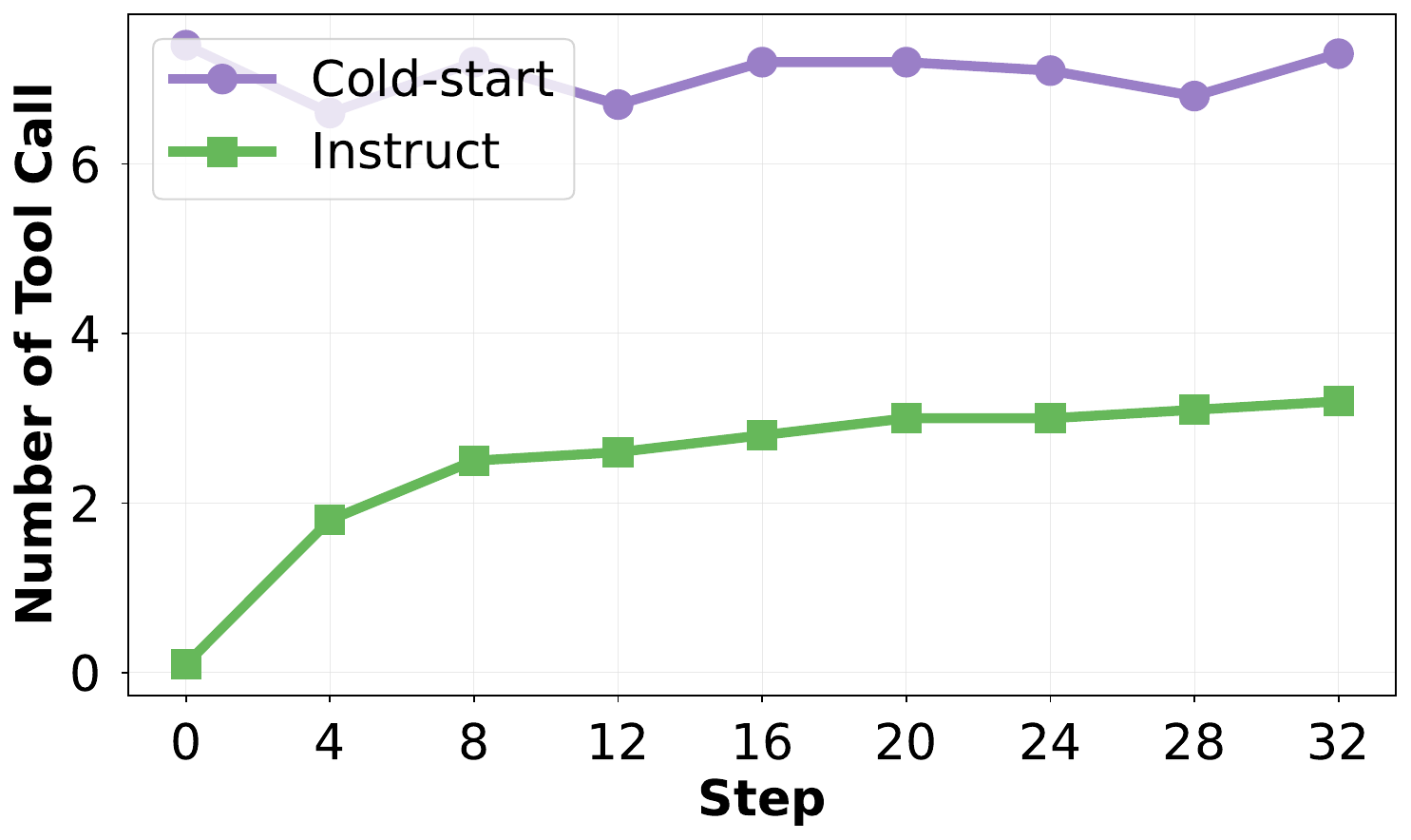}}
    \caption{Comparing direct RL training of Qwen-2.5-instruct-32B with RL training after an RFT cold start.}
    \label{fig:coldstart}
\end{figure}

\paragraph{With/without cold start} To evaluate the efficacy of our RFT cold-start strategy, we compare it against a direct RL training baseline, with results presented in Figure~\ref{fig:coldstart}. We observe that while the direct RL approach exhibits a larger increase in Pass@1 accuracy, the final converged performance of the model that underwent an RFT cold start is significantly superior. This performance gap is also reflected in the models' tool usage patterns. The tool call count for the cold-started model remains high and stable throughout RL training, whereas the tool call count for the direct RL model, despite a steady increase, remains substantially lower, indicating an inability to master long-horizon reasoning. Critically, the performance disparity between the cold-started WebSailor and the direct RL model is much wider on the BrowseComp-en. This suggests that without an RFT cold start, it is extremely difficult for a model to acquire the sophisticated reasoning patterns—often found only in powerful LRMs—through self-exploration alone. The cold start is essential for bootstrapping the model with these complex strategies, which are necessary to solve exceptionally challenging tasks.

\subsection{Limitations and Future Work} First, our decision to filter training trajectories to under 32k tokens, while pragmatic, may cap the model's ability to tackle even more complex problems. Our analysis of failed cases reveals that many errors stem from exceeding the context limit, and we observe that performance can degrade as inference length increases. Furthermore, WebSailor can exhibit a tendency for "over-thinking", applying multi-step tool calls even to seemingly simple questions. However, this is not a clear-cut drawback; our qualitative analysis suggests that in many such instances, the agent is not aimlessly exploring but performing cross-verification, using different information sources to validate an initial finding. Finally, on the training front, our RL process is limited to 50 steps. This is primarily due to the inherent inefficiency of the synchronous RL framework; even with the optimizations from DUPO, the training speed remains a bottleneck. Future work will focus on migrating to an asynchronous training framework to improve efficiency and enable more extensive RL training.

\section{Related Work}

\paragraph{Information seeking benchmarks}The landscape of information-seeking benchmarks has evolved from tasks with easily reducible uncertainty to those demanding complex, non-linear reasoning~\citep{webwalker}. Early datasets such as NQ~\citep{kwiatkowski2019natural}, TriviaQA~\citep{joshi2017triviaqa}, and multi-hop variants like HotpotQA~\citep{yang2018hotpotqa} and Musique~\citep{trivedi2022musique} represent problems where solutions can often be found through a structured sequence of queries or even from a model's parametric knowledge alone. More recent benchmarks have raised the complexity. GAIA~\citep{mialon2023gaia}, while a generalist multimodal benchmark, introduces information-seeking challenges that approach complex multi-hop QA. Similarly, Xbench-DeepSearch~\citep{xbench} specifically targets agents' deep search and tool-use capabilities through professionally annotated, dynamic tasks. At the apex of this evolution lie benchmarks like BrowseComp-en/zh~\citep{bc_en, bc_zh}, which embody the Level 3 complexity central to our work. These tasks are characterized by intricately coupled entities and deliberate information obfuscation, creating high initial uncertainty that is exceptionally difficult to reduce. Success on BrowseComp requires the kind of sophisticated, non-linear exploration and synthesis that defines superhuman reasoning, making it the ideal proving ground for advanced web agents.

\paragraph{Web agents}The development of autonomous web agents has witnessed significant progress from both proprietary and open-source communities~\citep{evolvesearch}. Proprietary systems like DeepResearch~\citep{dr}, Doubao with Deep Think~\citep{doubao}, and Grok-3~\citep{grok} have demonstrated superhuman performance in complex web navigation and information synthesis tasks, but their internal architectures and training methodologies remain opaque, impeding collaborative research. In contrast, open-source projects such as WebDancer~\citep{wu2025webdancer}, WebThinker~\citep{Li2025webthinker}, and R1-Searcher~\citep{song2025r1}, adopting the ReAct framework~\citep{yao2023react}, have made strides in simpler tasks yet face a substantial performance gap in benchmarks requiring sophisticated non-linear reasoning. In terms of training methodologies, the foundational principles of training dynamics outlined by ~\citep{xu2019frequency,xu2024overview,xu2025overview} offer crucial insights, informing hyperparameter tuning and algorithmic design by highlighting the decisive impact of the training approach on models' generalization ability. However, while many studies use Supervised Fine-Tuning (SFT) following the ReAct paradigm, pure SFT agents struggle with generalization in adaptive contexts~\citep{zheng2025deepresearch, zhang2025nemotron}. Reinforcement learning-based methods~\cite{song2025r1,zheng2025deepresearch} hold promise for advanced search strategies via learned exploration policies but encounter challenges in training stability and sample efficiency. 

\section{Conclusion}

In this work, we propose WebSailor. From the perspective of uncertainty reduction in information seeking, we analyze why previous open-source web agents have not reached the level of proprietary systems. Our contributions span from QA construction, comprehensive training data synthesis, RFT cold start, to improved efficiency in RL algorithms, leading to a full agentic post-training pipeline. WebSailor demonstrates strong performance on both simple and complex information seeking benchmarks, exhibiting reasoning and tool-use capabilities that surpass human levels.

We believe that the key to agentic post-training lies in further defining more complex tasks with higher uncertainty, as well as achieving more effective and efficient RL training. In the future, we will continue to explore how to further enhance agent capabilities based on open-source models, not only in the domain of information seeking, but also in pursuing the goal of general “superhuman” performance across more dimensions.

\clearpage
\appendix
\section{Experimental Details}
\label{apx:exp}
\subsection{Tools}
\label{apx:tool}
WebSailor uses two types of tools, search and visit:
\begin{itemize}
    \item \textbf{Search} is used to access the Google search engine for information retrieval. The parameters of Search are the search queries. It allows searching multiple queries simultaneously and returns the top-10 results for each query. Each result contains a title, a snippet, and the corresponding URL.
    \item \textbf{Visit} is used to access specific web pages. The input consists of several web pages and their corresponding visit goals, with each page having a dedicated goal. First, Jina~\citep{jina} is used to retrieve the full content of the web page, and then a summary model extracts relevant information based on the goal. In this paper, we use Qwen-2.5-72B as the summary model.
\end{itemize}

\subsection{QA Construction}
\label{apx:qa}
Our QA is constructed by sampling a subgraph from a graph. Each graph is generated by performing a random walk starting from a rare entity. The nodes in the graph represent entities, and the edges represent the relationships between entities. The general process for constructing the graph is as follows:
\begin{enumerate}
    \item We use Wikidata's SPARQL service to obtain rare entities based on certain database rules.
    \item The features of the initial node are obtained using the search and visit tools, and the initial node is set as the expansion node.
    \item Some related entiti es are obtained based on the features of the expansion node, and then we get their features.
    \item With a certain probability, we either set a new related entity as the next expansion node, or select a node from the previous nodes.
    \item Repeat steps 3 and 4 until the number of edges in the graph reaches a predefined value.
\end{enumerate}

\subsection{ReAct Trajectories}
Our ReAct framework is implemented through Qwen-Agent~\footnote{\url{https://github.com/QwenLM/Qwen-Agent/}}, and we limit the number of tool calls to no more than 30.
 A complete trajectory follows the format below:

\begin{tcolorbox}[title=Case Trajectory]
<think> thinking process here </think>\\
<tool\_call>\\
{"name": "tool name here", "arguments": {"parameter name here": parameter value here, "another parameter name here": another parameter value here, ...}}\\
</tool\_call>\\
<tool\_response>\\
tool\_response here\\
</tool\_response>\\
(more thinking processes, tool calls and tool responses here)\\
<think> thinking process here </think>\\
<answer> answer here </answer>\\
\end{tcolorbox}

\subsection{Training Details}
We use Megatron~\citep{shoeybi2019megatron} for SFT and verl~\citep{sheng2025hybridflow} for RL training. For SFT, we use a batch size of 32, learning rate of 5e-6 with a minimum of 1e-10, warmup plus cosine decay schedule, and a weight decay of 0.1. For RL training, the rollout number in a group is 8, the temperature is 1.0, $top_p=1.0$, the batch size is 128, the mini batch size is 32, and the learning rate is 1e-6.

\subsection{Case Study}
\begin{tcolorbox}[breakable,title=Case Trajectory in BrowseComp-en]
\textcolor{black}{\textbf{Question:} There is a software developer who, sometime after 2010 but before 2023, claimed to have designed and created a solar-powered fridge. They were known to live a rustic lifestyle sometime after 2012 and before 2023 and be located, in their words, "in a hole in the map". They have memories gleaned from a developer's conference in Edinburgh and were also known to enjoy caving. What was the make and model of the first computer purchased jointly with their father during the 1980s?}\\
\colorbox{gray!70}{%
    \parbox{\dimexpr\textwidth-2\fboxsep}{%
    \centering
        \textbf{Step 1}%
    }%
}
\textcolor{purple1}{\textbf{<think>} \\
The question requires identifying a software developer with a unique set of biographical details—someone who built a solar-powered fridge, lived a rustic lifestyle, described their location as “a hole in the map,” had memories from a developer’s conference in Edinburgh, and enjoyed caving. To make progress, I need to triangulate the individual's identity by searching for overlapping evidence of these distinctive traits online...\\
\textbf{</think>}}\\
\textcolor{black}{\textbf{<tool\_call>} \\ 
\{"name": "search", "arguments": \{"query": ["software developer solar-powered fridge rustic lifestyle", "software developer hole in the map caving"...]\}\}
\\
\textbf{</tool\_call>}}\\
\textcolor{black}{\textbf{<tool\_response>}\\
A Google search for 'software developer solar-powered fridge rustic lifestyle' found 10 results: \\
1. [Solar Fridge with Temperature Control and Monitoring System Using ...] (https://www.ijert.org/solar-fridge-with-temperature-control-and-monitoring-system-using-bluetooth-module) \\
Here we are designing a mini solar Bluetooth based refrigerator with temperature control with temperature monitoring system which is cheaper as well as eco- ... \\
2. [Engineer, Embedded Software - EnPhase Energy - Climatebase] (https://climatebase.org/job/61314385/engineer-embedded-software)
Date published: Jun 18, 2025\\
Enphase Energy is a global energy technology company and a leading provider of solar, battery, and electric vehicle charging products. \\
...\\
\textbf{</tool\_response>}}\\
\colorbox{gray!70}{%
    \parbox{\dimexpr\textwidth-2\fboxsep}{%
    \centering
        \textbf{Step 2}%
    }%
}
\textcolor{purple1}{\textbf{<think>} \\
The combination of unique biographical markers—such as a software developer with a history of building a solar-powered fridge, a self-described “hole in the map” rustic lifestyle, and a background in caving and developer conferences in Edinburgh—suggests that the right approach is to cross-reference these traits in more integrated search queries. By combining all these distinctive elements into multi-part searches... \\
\textbf{</think>}}\\
\textcolor{black}{\textbf{<tool\_call>} \\ 
\{"name": "search", "arguments": \{"query":  ["software developer rustic lifestyle caving conference Edinburgh", "solar-powered fridge caving conference Edinburgh rustic lifestyle"...]\}\}
\\
\textbf{</tool\_call>}}\\
\textcolor{black}{\textbf{<tool\_response>}\\
A Google search for 'software developer rustic lifestyle caving conference Edinburgh' found 3 results: \\
1. [Developer conferences in Edinburgh 2025 / 2026 - dev.events](https://dev.events/EU/GB/\\Edinburgh) \\
Missing: rustic lifestyle caving \\
2. [Joy - The Red Hand Files](https://www.theredhandfiles.com/joy/)\\
Some have likened joy to be a flame enshrined behind a glass lantern – no matter how hard the wind blows, the flicker stays intact. \\
...\\
\textbf{</tool\_response>}}\\
\colorbox{gray!70}{%
    \parbox{\dimexpr\textwidth-2\fboxsep}{%
    \centering
        \textbf{Step 3}%
    }%
}
\textcolor{purple1}{\textbf{<think>} \\
The phrase “hole in the map” is highly distinctive and likely to be quoted or paraphrased by the individual in question or in descriptions about them. By searching for combinations involving “software developer solar-powered fridge” and “hole in the map,” I’m aiming to surface direct references—such as interviews, blog posts, or forum profiles—where the person has described their lifestyle and technical projects in their own words... \\
\textbf{</think>}} \\
\textcolor{black}{\textbf{<tool\_call>} \\ 
\{"name": "search", "arguments": \{"query":   ["software developer solar-powered fridge \"hole in the map\"", "software developer \"hole in the map\" rustic lifestyle"...]\}\}
\\
\textbf{</tool\_call>}}\\
\textcolor{black}{\textbf{<tool\_response>}\\
A Google search for 'software developer solar-powered fridge "hole in the map"' found 5 results: \\
1. [[PDF] Observing With The Green Bank Telescope - Science](https://www.gb.nrao.edu/~glangsto/GBTog.pdf)
Date published: Oct 11, 2011 \\
the largest “hole” in the map is approximately 0.3 x map radius. However it is useful for focal-plane arrays. Syntax: Daisy(location, map ... \\
2. [[PDF] Larissa Fergus.pdf - VU Research Repository](https://vuir.vu.edu.au/30098/1/Larissa\%2\\
0Fergus.pdf)\\
I can see a tiny hole in the map, between contours, on the plain west of the mantelpiece. I lift my goggles and look over at the corresponding place on the\\
...\\
\textbf{</tool\_response>}}\\
\colorbox{gray!70}{%
    \parbox{\dimexpr\textwidth-2\fboxsep}{%
    \centering
        \textbf{Step 4}%
    }%
}
\textcolor{purple1}{\textbf{<think>} \\
The LWN.net article surfaced in the search results with a direct quote about living “in a hole in the map,” which matches the unique phrasing from the question and is strongly associated with a rustic lifestyle. This suggests the article may contain a first-person account or interview with the software developer in question, potentially revealing their identity and offering biographical details such as their interests in caving, solar-powered projects, and perhaps even recollections about early computing experiences... \\
\textbf{</think>}} \\
\textcolor{black}{\textbf{<tool\_call>}\\
\{"name": "visit", "arguments": \{"url": ["https://lwn.net/Articles/672352"], "goal": "Identify the software developer, their claim to have designed a solar-powered fridge, their rustic lifestyle mention, and any details about early computer purchases"\}\}\\
\textbf{</tool\_call>}}\\
\textcolor{black}{\textbf{<tool\_response>}\\
The useful information in https://lwn.net/Articles/672352/ for user goal Identify the software developer, their claim to have designed a solar-powered fridge, their rustic lifestyle mention, and any details about early computer purchases as follows: 
...\\
\textbf{</tool\_response>}}\\
\colorbox{gray!70}{%
    \parbox{\dimexpr\textwidth-2\fboxsep}{%
    \centering
        \textbf{Step 5}%
    }%
}
\textcolor{purple1}{\textbf{<think>} \\
The interview on LWN.net provides strong evidence that Joey Hess is the software developer in question, detailing his rustic lifestyle “in a hole in the map” and his early computing experiences. However, while there’s mention of solar panels and off-grid living, there’s no explicit confirmation yet that he claimed to have designed and created a solar-powered fridge—an important detail for fully verifying his identity as the subject of the question...\\
\textbf{</think>}} \\
\textcolor{black}{\textbf{<tool\_call>} \\ 
\{"name": "search", "arguments": \{"query": ["Joey Hess solar-powered fridge", "Joey Hess solar powered fridge created", "Joey Hess designed solar-powered fridge", "Joey Hess fridge solar"]\}\}
\\
\textbf{</tool\_call>}}\\
\textcolor{black}{\textbf{<tool\_response>}\\
A Google search for 'Joey Hess solar-powered fridge' found 9 results:\\
1. [Off-Grid, Solar-Powered, Zero-Battery Refrigerator](https://www.notechmagazine.com/2019/0\\
9/off-grid-solar-powered-zero-battery-refrigerator.html)\\
Date published: Sep 1, 2019 \\
Joey Hess has designed, built and tested an off-grid, solar powered fridge, with no battery bank. Using an inexpensive chest freezer with a few modifications. \\
2. [Fridge0: an open source hardware solar fridge with no battery ...](https://boingboing.net/2019/09/05/frozen-sunshine.html)
Date published: Sep 5, 2019\\
Joey Hess designed the first Fridge0 a year ago: it uses a standard chest freezer with added thermal mass, a simple controller, ...\\
...\\
\textbf{</tool\_response>}}\\
\colorbox{gray!70}{%
    \parbox{\dimexpr\textwidth-2\fboxsep}{%
    \centering
        \textbf{Step 6}%
    }%
}
\textcolor{purple1}{\textbf{<think>} \\
To fully verify that Joey Hess matches all the biographical clues in the question, I need to establish whether he has personal memories of a developer's conference in Edinburgh and is also known for enjoying caving. These two details are particularly distinctive and would help confirm his identity as the subject of the question, especially since I already have strong evidence about his rustic lifestyle and solar-powered fridge...\\
\textbf{</think>}} \\
\textcolor{black}{\textbf{<tool\_call>} \\ 
\{"name": "search", "arguments": \{"query":  ["Joey Hess developer's conference Edinburgh memories", "Joey Hess caving"]\}\}
\\
\textbf{</tool\_call>}}\\
\textcolor{black}{\textbf{<tool\_response>}\\
A Google search for 'Joey Hess developer's conference Edinburgh memories' found 9 results:\\
1. [Leaving Debian With Joey Hess - CoRecursive Podcast](https://corecursive.com/leaving-debian/)
Date published: Nov 2, 2021 \\
Joey Hess is my guest, and he dedicated significant parts of his adult life to working on Debian. He's going to share what that was like. \\
2. [Interviews with FLOSS developers: Joey Hess | i.am.z.grim.shell](https://zgrimshell.github.io/posts/interviews-with-floss-developers-joey-hess.html)
Date published: Feb 28, 2015\\
me: Some memorable moments from Debian conferences? joeyh: There are so many! Picnicing on berries and tamales at the Portland farmer's market ...\\
...\\
\textbf{</tool\_response>}}\\
\colorbox{gray!70}{%
    \parbox{\dimexpr\textwidth-2\fboxsep}{%
    \centering
        \textbf{Step 7}%
    }%
}
\textcolor{purple1}{\textbf{<think>} \\
To confidently establish whether Joey Hess has personal memories of a developer's conference in Edinburgh—an important detail for confirming he matches all the unique clues in the question—I need direct evidence of his experiences or references to such an event. The interview linked at zgrimshell.github.io specifically asks Joey Hess about memorable moments from Debian conferences... \\
\textbf{</think>}} \\
\textcolor{black}{\textbf{<tool\_call>}\\
\{"name": "visit", "arguments": \{"url": ["https://zgrimshell.github.io/posts/interviews-with-floss-developers-joey-hess.html"], "goal": "Find mention of Edinburgh and conference memories"\}\}\\
\textbf{</tool\_call>}}\\
\textcolor{black}{\textbf{<tool\_response>}\\
The useful information in https://zgrimshell.github.io/posts/interviews-with-floss-developers-joey-hess.html for user goal Find mention of Edinburgh and conference memories as follows:...\\
\textbf{</tool\_response>}}\\
\colorbox{gray!70}{%
    \parbox{\dimexpr\textwidth-2\fboxsep}{%
    \centering
        \textbf{Step 8}%
    }%
}
\textcolor{purple1}{\textbf{<think>} \\
While I have strong evidence linking Joey Hess to the rustic lifestyle, solar-powered fridge, and developer conference memories, the detail about caving is still unconfirmed. Since "caving" could be a metaphor or refer to an actual hobby, I want to directly verify whether Joey Hess has publicly mentioned caving or spelunking as an interest or activity...\\
\textbf{</think>}} \\
\textcolor{black}{\textbf{<tool\_call>} \\ 
\{"name": "search", "arguments": \{"query":   ["Joey Hess caving", "Joey Hess cave", "Joey Hess spelunking"]\}\}
\\
\textbf{</tool\_call>}}\\
\textcolor{black}{\textbf{<tool\_response>}\\
A Google search for 'Joey Hess caving' found 9 results:\\
1. [Joe Hess on X: ""So come out of your cave walking on your hands ...](https://twitter.com/RunningJoeInJax/status/104570999537799168) \\
"So come out of your cave walking on your hands, and see the world hanging upside down..." \#TheCave \#Muford\&Sons. 3:10 PM · Aug 19, 2011. \\
2. [Joe Hess (@jhessacx) • Instagram photos and videos](https://www.instagram.com/jhessacx/)\\
Day 5 - Kingdom Church and 3 baptisms in the Danube River. The testimonies of the baptism candidates were heartfelt to witness and the after baptism prayers ...
\\
...\\
\textbf{</tool\_response>}}\\
\colorbox{gray!70}{%
    \parbox{\dimexpr\textwidth-2\fboxsep}{%
    \centering
        \textbf{Step 9}%
    }%
}
\textcolor{purple1}{\textbf{<think>} \\
To determine the make and model of the first computer purchased jointly with his father, I need to locate a source where Joey Hess has directly discussed this formative experience, ideally in his own words. The blog post at the provided URL is likely to contain personal anecdotes and technical details about his early computing history, which is precisely the information required to answer the question... \\
\textbf{</think>}} \\
\textcolor{black}{\textbf{<tool\_call>}\\
\{"name": "visit", "arguments": \{"url": ["http://joeyh.name/blog/entry/atari"],, "goal": "Find the make and model of the first computer purchased with his father in the 1980s"\}\}\\
\textbf{</tool\_call>}}\\
\textcolor{black}{\textbf{<tool\_response>}\\
The useful information in http://joeyh.name/blog/entry/atari/ for user goal Find the make and model of the first computer purchased with his father in the 1980s as follows: ...\\
\textbf{</tool\_response>}}\\
\colorbox{gray!70}{%
    \parbox{\dimexpr\textwidth-2\fboxsep}{%
    \centering
        \textbf{Step 10}%
    }%
}
\textcolor{purple1}{\textbf{<think>} \\
To arrive at the answer "Atari 130XE", I first needed to identify the software developer described in the question by piecing together several unique biographical clues: designing a solar-powered fridge, living a rustic lifestyle "in a hole in the map", having memories from a developer's conference in Edinburgh, and enjoying caving. My searches led me to Joey Hess, whose interviews and blog posts explicitly confirm his off-grid living, the creation of the "fridge0" solar-powered fridge, and fond recollections of Debian conferences—including an event in Edinburgh. To answer the core question about the first computer purchased with his father in the 1980s, I located Joey's blog entry detailing that it was an Atari 130XE, bought jointly with his dad around 1986. This direct evidence from his own writing allowed me to confidently conclude that the Atari 130XE was the correct answer. \\
\textbf{</think>}} \\
\textcolor{red}{\textbf{<answer>}The first computer they bought together was an Atari 130XE.
\textbf{<answer>}} \textcolor{black}{\checkmark}\\
\end{tcolorbox}

\clearpage
\bibliography{biblio}

\begin{thebibliography}{60}
\providecommand{\natexlab}[1]{#1}
\providecommand{\url}[1]{\texttt{#1}}
\expandafter\ifx\csname urlstyle\endcsname\relax
  \providecommand{\doi}[1]{doi: #1}\else
  \providecommand{\doi}{doi: \begingroup \urlstyle{rm}\Url}\fi

\bibitem[Amodei et~al.(2016)Amodei, Olah, Steinhardt, Christiano, Schulman, and Man{\'e}]{amodei2016concrete}
Dario Amodei, Chris Olah, Jacob Steinhardt, Paul Christiano, John Schulman, and Dan Man{\'e}.
\newblock Concrete problems in ai safety.
\newblock \emph{arXiv preprint arXiv:1606.06565}, 2016.

\bibitem[Chen et~al.(2023)Chen, Shu, Shareghi, Collier, Narasimhan, and Yao]{chen2023fireact}
Baian Chen, Chang Shu, Ehsan Shareghi, Nigel Collier, Karthik Narasimhan, and Shunyu Yao.
\newblock Fireact: Toward language agent fine-tuning.
\newblock \emph{arXiv preprint arXiv:2310.05915}, 2023.

\bibitem[Chen et~al.(2025)Chen, Tu, Wang, Liu, Tang, Du, Zhou, and Xie]{chen2025sft}
Hardy Chen, Haoqin Tu, Fali Wang, Hui Liu, Xianfeng Tang, Xinya Du, Yuyin Zhou, and Cihang Xie.
\newblock Sft or rl? an early investigation into training r1-like reasoning large vision-language models.
\newblock \emph{arXiv preprint arXiv:2504.11468}, 2025.

\bibitem[Chen et~al.(2021)Chen, Tworek, Jun, Yuan, Pinto, Kaplan, Edwards, Burda, Joseph, Brockman, et~al.]{chen2021evaluating}
Mark Chen, Jerry Tworek, Heewoo Jun, Qiming Yuan, Henrique Ponde De~Oliveira Pinto, Jared Kaplan, Harri Edwards, Yuri Burda, Nicholas Joseph, Greg Brockman, et~al.
\newblock Evaluating large language models trained on code.
\newblock \emph{arXiv preprint arXiv:2107.03374}, 2021.

\bibitem[Chu et~al.(2025)Chu, Zhai, Yang, Tong, Xie, Schuurmans, Le, Levine, and Ma]{chu2025sft}
Tianzhe Chu, Yuexiang Zhai, Jihan Yang, Shengbang Tong, Saining Xie, Dale Schuurmans, Quoc~V Le, Sergey Levine, and Yi~Ma.
\newblock Sft memorizes, rl generalizes: A comparative study of foundation model post-training.
\newblock \emph{arXiv preprint arXiv:2501.17161}, 2025.

\bibitem[Doubao(2025)]{doubao}
ByteDance Doubao.
\newblock Doubao, 2025.
\newblock URL \url{http://www.doubao.com/}.

\bibitem[Google(2020)]{compge}
Google.
\newblock Measuring compositional generalization, 2020.
\newblock URL \url{https://research.google/blog/measuring-compositional-generalization/}.

\bibitem[Guo et~al.(2025)Guo, Yang, Zhang, Song, Zhang, Xu, Zhu, Ma, Wang, Bi, et~al.]{r1}
Daya Guo, Dejian Yang, Haowei Zhang, Junxiao Song, Ruoyu Zhang, Runxin Xu, Qihao Zhu, Shirong Ma, Peiyi Wang, Xiao Bi, et~al.
\newblock {DeepSeek-R1}: Incentivizing reasoning capability in {LLMs} via reinforcement learning.
\newblock \emph{arXiv preprint arXiv:2501.12948}, 2025.

\bibitem[Ho et~al.(2020)Ho, Nguyen, Sugawara, and Aizawa]{ho2020constructingmultihopqadataset}
Xanh Ho, Anh-Khoa~Duong Nguyen, Saku Sugawara, and Akiko Aizawa.
\newblock Constructing a multi-hop qa dataset for comprehensive evaluation of reasoning steps, 2020.
\newblock URL \url{https://arxiv.org/abs/2011.01060}.

\bibitem[Hu et~al.(2025)Hu, Zhang, Han, Jiang, Zhang, and Shum]{hu2025open}
Jingcheng Hu, Yinmin Zhang, Qi~Han, Daxin Jiang, Xiangyu Zhang, and Heung-Yeung Shum.
\newblock Open-reasoner-zero: An open source approach to scaling up reinforcement learning on the base model.
\newblock \emph{arXiv preprint arXiv:2503.24290}, 2025.

\bibitem[Huang et~al.(2023)Huang, Song, Wang, Zhao, Chen, Juefei-Xu, and Ma]{huang2023look}
Yuheng Huang, Jiayang Song, Zhijie Wang, Shengming Zhao, Huaming Chen, Felix Juefei-Xu, and Lei Ma.
\newblock Look before you leap: An exploratory study of uncertainty measurement for large language models.
\newblock \emph{arXiv preprint arXiv:2307.10236}, 2023.

\bibitem[Jin et~al.(2025)Jin, Zeng, Yue, Yoon, Arik, Wang, Zamani, and Han]{jin2025search}
Bowen Jin, Hansi Zeng, Zhenrui Yue, Jinsung Yoon, Sercan Arik, Dong Wang, Hamed Zamani, and Jiawei Han.
\newblock Search-r1: Training llms to reason and leverage search engines with reinforcement learning.
\newblock \emph{arXiv preprint arXiv:2503.09516}, 2025.

\bibitem[Jina.ai(2025)]{jina}
Jina.ai.
\newblock Jina, 2025.
\newblock URL \url{https://jina.ai/}.

\bibitem[Joshi et~al.(2017)Joshi, Choi, Weld, and Zettlemoyer]{joshi2017triviaqa}
Mandar Joshi, Eunsol Choi, Daniel~S Weld, and Luke Zettlemoyer.
\newblock Triviaqa: A large scale distantly supervised challenge dataset for reading comprehension.
\newblock \emph{arXiv preprint arXiv:1705.03551}, 2017.

\bibitem[Jurado et~al.(2015)Jurado, Ludvigson, and Ng]{jurado2015measuring}
Kyle Jurado, Sydney~C Ludvigson, and Serena Ng.
\newblock Measuring uncertainty.
\newblock \emph{American Economic Review}, 105\penalty0 (3):\penalty0 1177--1216, 2015.

\bibitem[Kapoor et~al.(2024)Kapoor, Gruver, Roberts, Collins, Pal, Bhatt, Weller, Dooley, Goldblum, and Wilson]{kapoor2024large}
Sanyam Kapoor, Nate Gruver, Manley Roberts, Katherine Collins, Arka Pal, Umang Bhatt, Adrian Weller, Samuel Dooley, Micah Goldblum, and Andrew~Gordon Wilson.
\newblock Large language models must be taught to know what they don't know.
\newblock \emph{arXiv preprint arXiv:2406.08391}, 2024.

\bibitem[Kwiatkowski et~al.(2019)Kwiatkowski, Palomaki, Redfield, Collins, Parikh, Alberti, Epstein, Polosukhin, Devlin, Lee, et~al.]{kwiatkowski2019natural}
Tom Kwiatkowski, Jennimaria Palomaki, Olivia Redfield, Michael Collins, Ankur Parikh, Chris Alberti, Danielle Epstein, Illia Polosukhin, Jacob Devlin, Kenton Lee, et~al.
\newblock Natural questions: a benchmark for question answering research.
\newblock \emph{Transactions of the Association for Computational Linguistics}, 7:\penalty0 453--466, 2019.

\bibitem[Li et~al.(2025{\natexlab{a}})Li, Zhang, Jiang, Xie, Huang, Wang, and Cheng]{li2025lara}
Kuan Li, Liwen Zhang, Yong Jiang, Pengjun Xie, Fei Huang, Shuai Wang, and Minhao Cheng.
\newblock Lara: Benchmarking retrieval-augmented generation and long-context llms--no silver bullet for lc or rag routing.
\newblock \emph{arXiv preprint arXiv:2502.09977}, 2025{\natexlab{a}}.

\bibitem[Li et~al.(2025{\natexlab{b}})Li, Dong, Jin, Zhang, Zhou, Zhu, Zhang, and Dou]{li2025search}
Xiaoxi Li, Guanting Dong, Jiajie Jin, Yuyao Zhang, Yujia Zhou, Yutao Zhu, Peitian Zhang, and Zhicheng Dou.
\newblock Search-o1: Agentic search-enhanced large reasoning models.
\newblock \emph{arXiv preprint arXiv:2501.05366}, 2025{\natexlab{b}}.

\bibitem[Li et~al.(2025{\natexlab{c}})Li, Jin, Dong, Qian, Zhu, Wu, Wen, and Dou]{Li2025webthinker}
Xiaoxi Li, Jiajie Jin, Guanting Dong, Hongjin Qian, Yutao Zhu, Yongkang Wu, Ji{-}Rong Wen, and Zhicheng Dou.
\newblock Webthinker: Empowering large reasoning models with deep research capability.
\newblock \emph{CoRR}, abs/2504.21776, 2025{\natexlab{c}}.
\newblock \doi{10.48550/ARXIV.2504.21776}.
\newblock URL \url{https://doi.org/10.48550/arXiv.2504.21776}.

\bibitem[Liu et~al.(2024)Liu, Yang, Huang, Zhang, Huang, Wei, Deng, Sun, and Zhang]{DBLP:conf/coling/LiuYHZHWDSZ24}
Yuxuan Liu, Tianchi Yang, Shaohan Huang, Zihan Zhang, Haizhen Huang, Furu Wei, Weiwei Deng, Feng Sun, and Qi~Zhang.
\newblock Calibrating llm-based evaluator.
\newblock In Nicoletta Calzolari, Min{-}Yen Kan, V{\'{e}}ronique Hoste, Alessandro Lenci, Sakriani Sakti, and Nianwen Xue (eds.), \emph{Proceedings of the 2024 Joint International Conference on Computational Linguistics, Language Resources and Evaluation, {LREC/COLING} 2024, 20-25 May, 2024, Torino, Italy}, pp.\  2638--2656. {ELRA} and {ICCL}, 2024.
\newblock URL \url{https://aclanthology.org/2024.lrec-main.237}.

\bibitem[Mialon et~al.(2023)Mialon, Fourrier, Wolf, LeCun, and Scialom]{mialon2023gaia}
Gr{\'e}goire Mialon, Cl{\'e}mentine Fourrier, Thomas Wolf, Yann LeCun, and Thomas Scialom.
\newblock Gaia: a benchmark for general ai assistants.
\newblock In \emph{The Twelfth International Conference on Learning Representations}, 2023.

\bibitem[{OpenAI}(2024)]{gpt4o}
{OpenAI}.
\newblock Hello {GPT-4o}, 2024.
\newblock URL \url{https://openai.com/index/hello-gpt-4o/}.

\bibitem[OpenAI(2025{\natexlab{a}})]{dr}
OpenAI.
\newblock Deep research system card, 2025{\natexlab{a}}.
\newblock URL \url{https://cdn.openai.com/deep-research-system-card.pdf}.

\bibitem[OpenAI(2025{\natexlab{b}})]{gpt4.1}
OpenAI.
\newblock Introducing openai gpt-4.1, 2025{\natexlab{b}}.
\newblock URL \url{https://openai.com/index/gpt-4-1/}.

\bibitem[OpenAI(2025{\natexlab{c}})]{o3}
OpenAI.
\newblock Introducing openai o3 and o4-mini, 2025{\natexlab{c}}.
\newblock URL \url{https://openai.com/index/introducing-o3-and-o4-mini/}.

\bibitem[OpenAI(2025{\natexlab{d}})]{simpleqa}
OpenAI.
\newblock Introducing simpleqa, 2025{\natexlab{d}}.
\newblock URL \url{https://openai.com/index/introducing-simpleqa/}.

\bibitem[{Qwen~Team}(2025)]{qwq32b}
{Qwen~Team}.
\newblock {QwQ-32B}: Embracing the power of reinforcement learning, March 2025.
\newblock URL \url{https://qwenlm.github.io/blog/qwq-32b/}.

\bibitem[Shao et~al.(2024)Shao, Wang, Zhu, Xu, Song, Bi, Zhang, Zhang, Li, Wu, et~al.]{shao2024deepseekmath}
Zhihong Shao, Peiyi Wang, Qihao Zhu, Runxin Xu, Junxiao Song, Xiao Bi, Haowei Zhang, Mingchuan Zhang, YK~Li, Y~Wu, et~al.
\newblock Deepseekmath: Pushing the limits of mathematical reasoning in open language models.
\newblock \emph{arXiv preprint arXiv:2402.03300}, 2024.

\bibitem[Sheng et~al.(2025)Sheng, Zhang, Ye, Wu, Zhang, Zhang, Peng, Lin, and Wu]{sheng2025hybridflow}
Guangming Sheng, Chi Zhang, Zilingfeng Ye, Xibin Wu, Wang Zhang, Ru~Zhang, Yanghua Peng, Haibin Lin, and Chuan Wu.
\newblock Hybridflow: A flexible and efficient rlhf framework.
\newblock In \emph{Proceedings of the Twentieth European Conference on Computer Systems}, pp.\  1279--1297, 2025.

\bibitem[Shoeybi et~al.(2019)Shoeybi, Patwary, Puri, LeGresley, Casper, and Catanzaro]{shoeybi2019megatron}
Mohammad Shoeybi, Mostofa Patwary, Raul Puri, Patrick LeGresley, Jared Casper, and Bryan Catanzaro.
\newblock Megatron-lm: Training multi-billion parameter language models using model parallelism.
\newblock \emph{arXiv preprint arXiv:1909.08053}, 2019.

\bibitem[Song et~al.(2025)Song, Jiang, Min, Chen, Chen, Zhao, Fang, and Wen]{song2025r1}
Huatong Song, Jinhao Jiang, Yingqian Min, Jie Chen, Zhipeng Chen, Wayne~Xin Zhao, Lei Fang, and Ji-Rong Wen.
\newblock R1-searcher: Incentivizing the search capability in llms via reinforcement learning.
\newblock \emph{arXiv preprint arXiv:2503.05592}, 2025.

\bibitem[Sun et~al.(2024)Sun, Huang, and Pompili]{sun2024llm}
Chuanneng Sun, Songjun Huang, and Dario Pompili.
\newblock Llm-based multi-agent reinforcement learning: Current and future directions.
\newblock \emph{arXiv preprint arXiv:2405.11106}, 2024.

\bibitem[Sun et~al.(2025)Sun, Zhou, Wang, Li, Dziri, and Song]{sun2025climbing}
Yiyou Sun, Georgia Zhou, Hao Wang, Dacheng Li, Nouha Dziri, and Dawn Song.
\newblock Climbing the ladder of reasoning: What llms can-and still can't-solve after sft?
\newblock \emph{arXiv preprint arXiv:2504.11741}, 2025.

\bibitem[Swamy et~al.(2025)Swamy, Choudhury, Sun, Wu, and Bagnell]{swamy2025all}
Gokul Swamy, Sanjiban Choudhury, Wen Sun, Zhiwei~Steven Wu, and J~Andrew Bagnell.
\newblock All roads lead to likelihood: The value of reinforcement learning in fine-tuning.
\newblock \emph{arXiv preprint arXiv:2503.01067}, 2025.

\bibitem[Trivedi et~al.(2022)Trivedi, Balasubramanian, Khot, and Sabharwal]{trivedi2022musique}
Harsh Trivedi, Niranjan Balasubramanian, Tushar Khot, and Ashish Sabharwal.
\newblock Musique: Multihop questions via single-hop question composition.
\newblock \emph{Transactions of the Association for Computational Linguistics}, 10:\penalty0 539--554, 2022.

\bibitem[Wang et~al.(2024)Wang, Chen, Cheng, Liao, Zhang, Wu, Yu, Xu, Zhang, Luo, Li, Yang, Huang, and Li]{DBLP:conf/emnlp/WangCCL0WYXZLLY24}
Minzheng Wang, Longze Chen, Fu~Cheng, Shengyi Liao, Xinghua Zhang, Bingli Wu, Haiyang Yu, Nan Xu, Lei Zhang, Run Luo, Yunshui Li, Min Yang, Fei Huang, and Yongbin Li.
\newblock Leave no document behind: Benchmarking long-context llms with extended multi-doc {QA}.
\newblock In Yaser Al{-}Onaizan, Mohit Bansal, and Yun{-}Nung Chen (eds.), \emph{Proceedings of the 2024 Conference on Empirical Methods in Natural Language Processing, {EMNLP} 2024, Miami, FL, USA, November 12-16, 2024}, pp.\  5627--5646. Association for Computational Linguistics, 2024.
\newblock URL \url{https://aclanthology.org/2024.emnlp-main.322}.

\bibitem[Wei et~al.(2022)Wei, Wang, Schuurmans, Bosma, Xia, Chi, Le, Zhou, et~al.]{wei2022chain}
Jason Wei, Xuezhi Wang, Dale Schuurmans, Maarten Bosma, Fei Xia, Ed~Chi, Quoc~V Le, Denny Zhou, et~al.
\newblock Chain-of-thought prompting elicits reasoning in large language models.
\newblock \emph{Advances in neural information processing systems}, 35:\penalty0 24824--24837, 2022.

\bibitem[Wei et~al.(2024)Wei, Karina, Chung, Jiao, Papay, Glaese, Schulman, and Fedus]{wei2024measuring}
Jason Wei, Nguyen Karina, Hyung~Won Chung, Yunxin~Joy Jiao, Spencer Papay, Amelia Glaese, John Schulman, and William Fedus.
\newblock Measuring short-form factuality in large language models.
\newblock \emph{arXiv preprint arXiv:2411.04368}, 2024.

\bibitem[Wei et~al.(2025)Wei, Sun, Papay, McKinney, Han, Fulford, Chung, Passos, Fedus, and Glaese]{bc_en}
Jason Wei, Zhiqing Sun, Spencer Papay, Scott McKinney, Jeffrey Han, Isa Fulford, Hyung~Won Chung, Alex~Tachard Passos, William Fedus, and Amelia Glaese.
\newblock Browsecomp: A simple yet challenging benchmark for browsing agents.
\newblock \emph{arXiv preprint arXiv:2504.12516}, 2025.

\bibitem[Wiedemer et~al.(2023)Wiedemer, Mayilvahanan, Bethge, and Brendel]{wiedemer2023compositional}
Thadd{\"a}us Wiedemer, Prasanna Mayilvahanan, Matthias Bethge, and Wieland Brendel.
\newblock Compositional generalization from first principles.
\newblock \emph{Advances in Neural Information Processing Systems}, 36:\penalty0 6941--6960, 2023.

\bibitem[Wilson(1999)]{wilson1999models}
Tom~D Wilson.
\newblock Models in information behaviour research.
\newblock \emph{Journal of documentation}, 55\penalty0 (3):\penalty0 249--270, 1999.

\bibitem[Wu et~al.(2025{\natexlab{a}})Wu, Li, Fang, Yin, Zhang, Tao, Zhang, Xi, Jiang, Xie, et~al.]{wu2025webdancer}
Jialong Wu, Baixuan Li, Runnan Fang, Wenbiao Yin, Liwen Zhang, Zhengwei Tao, Dingchu Zhang, Zekun Xi, Yong Jiang, Pengjun Xie, et~al.
\newblock Webdancer: Towards autonomous information seeking agency.
\newblock \emph{arXiv preprint arXiv:2505.22648}, 2025{\natexlab{a}}.

\bibitem[Wu et~al.(2025{\natexlab{b}})Wu, Yin, Jiang, Wang, Xi, Fang, Zhang, He, Zhou, Xie, and Huang]{webwalker}
Jialong Wu, Wenbiao Yin, Yong Jiang, Zhenglin Wang, Zekun Xi, Runnan Fang, Linhai Zhang, Yulan He, Deyu Zhou, Pengjun Xie, and Fei Huang.
\newblock Webwalker: Benchmarking llms in web traversal, 2025{\natexlab{b}}.
\newblock URL \url{https://arxiv.org/abs/2501.07572}.

\bibitem[x.ai(2025)]{grok}
x.ai.
\newblock Grok 3 beta — the age of reasoning agents, 2025.
\newblock URL \url{https://x.ai/news/grok-3}.

\bibitem[Xbench-Team(2025)]{xbench}
Xbench-Team.
\newblock Xbench-deepsearch, 2025.
\newblock URL \url{https://xbench.org/agi/aisearch}.

\bibitem[Xu et~al.(2019)Xu, Zhang, Luo, Xiao, and Ma]{xu2019frequency}
Zhi-Qin~John Xu, Yaoyu Zhang, Tao Luo, Yanyang Xiao, and Zheng Ma.
\newblock Frequency principle: Fourier analysis sheds light on deep neural networks.
\newblock \emph{arXiv preprint arXiv:1901.06523}, 2019.

\bibitem[Xu et~al.(2024)Xu, Zhang, and Luo]{xu2024overview}
Zhi-Qin~John Xu, Yaoyu Zhang, and Tao Luo.
\newblock Overview frequency principle/spectral bias in deep learning.
\newblock \emph{Communications on Applied Mathematics and Computation}, pp.\  1--38, 2024.

\bibitem[Xu et~al.(2025)Xu, Zhang, and Zhou]{xu2025overview}
Zhi-Qin~John Xu, Yaoyu Zhang, and Zhangchen Zhou.
\newblock An overview of condensation phenomenon in deep learning.
\newblock \emph{arXiv preprint arXiv:2504.09484}, 2025.

\bibitem[Yang et~al.(2024)Yang, Yang, Zhang, Hui, Zheng, Yu, Li, Liu, Huang, Wei, et~al.]{qwen2.5}
An~Yang, Baosong Yang, Beichen Zhang, Binyuan Hui, Bo~Zheng, Bowen Yu, Chengyuan Li, Dayiheng Liu, Fei Huang, Haoran Wei, et~al.
\newblock Qwen2.5 technical report.
\newblock \emph{arXiv preprint arXiv:2412.15115}, 2024.

\bibitem[Yang et~al.(2018)Yang, Qi, Zhang, Bengio, Cohen, Salakhutdinov, and Manning]{yang2018hotpotqa}
Zhilin Yang, Peng Qi, Saizheng Zhang, Yoshua Bengio, William~W Cohen, Ruslan Salakhutdinov, and Christopher~D Manning.
\newblock Hotpotqa: A dataset for diverse, explainable multi-hop question answering.
\newblock \emph{arXiv preprint arXiv:1809.09600}, 2018.

\bibitem[Yao et~al.(2023)Yao, Zhao, Yu, Du, Shafran, Narasimhan, and Cao]{yao2023react}
Shunyu Yao, Jeffrey Zhao, Dian Yu, Nan Du, Izhak Shafran, Karthik Narasimhan, and Yuan Cao.
\newblock React: Synergizing reasoning and acting in language models.
\newblock In \emph{International Conference on Learning Representations (ICLR)}, 2023.

\bibitem[Ye et~al.(2025)Ye, Huang, Xiao, Chern, Xia, and Liu]{ye2025limo}
Yixin Ye, Zhen Huang, Yang Xiao, Ethan Chern, Shijie Xia, and Pengfei Liu.
\newblock Limo: Less is more for reasoning.
\newblock \emph{arXiv preprint arXiv:2502.03387}, 2025.

\bibitem[Yin et~al.(2025)Yin, Zhao, Wu, Ni, Zeng, Wang, Shi, Shao, Lyu, Wang, et~al.]{yin2025towards}
Huifeng Yin, Yu~Zhao, Minghao Wu, Xuanfan Ni, Bo~Zeng, Hao Wang, Tianqi Shi, Liangying Shao, Chenyang Lyu, Longyue Wang, et~al.
\newblock Towards widening the distillation bottleneck for reasoning models.
\newblock \emph{arXiv e-prints}, pp.\  arXiv--2503, 2025.

\bibitem[Yu et~al.(2025)Yu, Zhang, Zhu, Yuan, Zuo, Yue, Fan, Liu, Liu, Liu, et~al.]{yu2025dapo}
Qiying Yu, Zheng Zhang, Ruofei Zhu, Yufeng Yuan, Xiaochen Zuo, Yu~Yue, Tiantian Fan, Gaohong Liu, Lingjun Liu, Xin Liu, et~al.
\newblock Dapo: An open-source llm reinforcement learning system at scale.
\newblock \emph{arXiv preprint arXiv:2503.14476}, 2025.

\bibitem[Yue et~al.(2025)Yue, Chen, Lu, Zhao, Wang, Song, and Huang]{yue2025does}
Yang Yue, Zhiqi Chen, Rui Lu, Andrew Zhao, Zhaokai Wang, Shiji Song, and Gao Huang.
\newblock Does reinforcement learning really incentivize reasoning capacity in llms beyond the base model?
\newblock \emph{arXiv preprint arXiv:2504.13837}, 2025.

\bibitem[Zhang et~al.(2025{\natexlab{a}})Zhang, Zhao, Wu, Li, Yin, Zhang, Jiang, Li, Tu, Xie, and Huang]{evolvesearch}
Dingchu Zhang, Yida Zhao, Jialong Wu, Baixuan Li, Wenbiao Yin, Liwen Zhang, Yong Jiang, Yufeng Li, Kewei Tu, Pengjun Xie, and Fei Huang.
\newblock Evolvesearch: An iterative self-evolving search agent, 2025{\natexlab{a}}.
\newblock URL \url{https://arxiv.org/abs/2505.22501}.

\bibitem[Zhang et~al.(2025{\natexlab{b}})Zhang, Dong, Zhang, Kautz, Catanzaro, Tao, Wu, Yu, and Liu]{zhang2025nemotron}
Shaokun Zhang, Yi~Dong, Jieyu Zhang, Jan Kautz, Bryan Catanzaro, Andrew Tao, Qingyun Wu, Zhiding Yu, and Guilin Liu.
\newblock Nemotron-research-tool-n1: Tool-using language models with reinforced reasoning.
\newblock \emph{arXiv preprint arXiv:2505.00024}, 2025{\natexlab{b}}.

\bibitem[Zheng et~al.(2025)Zheng, Fu, Hu, Cai, Ye, Lu, and Liu]{zheng2025deepresearch}
Yuxiang Zheng, Dayuan Fu, Xiangkun Hu, Xiaojie Cai, Lyumanshan Ye, Pengrui Lu, and Pengfei Liu.
\newblock Deepresearcher: Scaling deep research via reinforcement learning in real-world environments, 2025.
\newblock URL \url{https://arxiv.org/abs/2504.03160}.

\bibitem[Zhou et~al.(2025)Zhou, Leon, Ying, Zhang, Shao, Ye, Chong, Jin, Xie, Cao, et~al.]{bc_zh}
Peilin Zhou, Bruce Leon, Xiang Ying, Can Zhang, Yifan Shao, Qichen Ye, Dading Chong, Zhiling Jin, Chenxuan Xie, Meng Cao, et~al.
\newblock Browsecomp-zh: Benchmarking web browsing ability of large language models in chinese.
\newblock \emph{arXiv preprint arXiv:2504.19314}, 2025.

\end{thebibliography}
\bibliographystyle{colm2024_conference}

\end{document}